\documentclass[10pt,journal,compsoc]{IEEEtran}

\ifCLASSOPTIONcompsoc
  \usepackage[nocompress]{cite}
\else
  \usepackage{cite}
\fi

\ifCLASSINFOpdf
\else
\fi

\usepackage[dvipsnames]{xcolor}
\usepackage{amsmath,amsfonts}
\usepackage{algorithm}
\usepackage{algorithmic}

\usepackage{array}
\usepackage[caption=false,font=normalsize,labelfont=sf,textfont=sf]{subfig}
\usepackage{textcomp}
\usepackage{stfloats}
\usepackage{url}
\usepackage{verbatim}
\usepackage{booktabs}
\usepackage{etoolbox}
\makeatletter
\patchcmd{\@makecaption}
  {\scshape}
  {}
  {}
  {}
\makeatletter
\patchcmd{\@makecaption}
  {\\}
  {.\ }
  {}
  {}
\makeatother

\usepackage{stfloats}
\usepackage{multirow}
\usepackage{amssymb}
\usepackage[mathscr]{eucal}
\usepackage{graphicx}
\usepackage{svg}
\usepackage{colortbl}  
\usepackage[switch]{lineno}
\begin{document}

\title{Low-Dimensional Gradient Helps Out-of-Distribution Detection}

\author{Yingwen Wu, Tao Li, Xinwen Cheng, Jie Yang~\IEEEmembership{Senior Member,~IEEE}, Xiaolin Huang~\IEEEmembership{Senior Member,~IEEE}
\thanks{Y. Wu, T. Li, X. Cheng, J. Yang and X. Huang are with the Department of Automation, Shanghai Jiao Tong University, 200240 Shanghai, P.R. China. Email: $\{$yingwen$\_$wu, li.tao, xinwencheng, jieyang, xiaolinhuang$\}$@sjtu.edu.cn
}
}

\markboth{submitted to IEEE Transactions on Pattern Analysis and Machine Intelligence}%
{Wu \MakeLowercase{\textit{et al.}}: Low Dimensional Gradient Helps OOD Detection}


\IEEEtitleabstractindextext{%
\begin{abstract}
Detecting out-of-distribution (OOD) samples is essential for ensuring the reliability of deep neural networks (DNNs) in real-world scenarios. While previous research has predominantly investigated the disparity between in-distribution (ID) and OOD data through forward information analysis, the discrepancy in parameter gradients during the backward process of DNNs has received insufficient attention. Existing studies on gradient disparities mainly focus on the utilization of gradient norms, neglecting the wealth of information embedded in gradient directions. To bridge this gap, in this paper, we conduct a comprehensive investigation into leveraging the entirety of gradient information for OOD detection. The primary challenge arises from the high dimensionality of gradients due to the large number of network parameters. To solve this problem, we propose performing linear dimension reduction on the gradient using a designated subspace that comprises principal components. This innovative technique enables us to obtain a low-dimensional representation of the gradient with minimal information loss. Subsequently, by integrating the reduced gradient with various existing detection score functions, our approach demonstrates superior performance across a wide range of detection tasks. For instance, on the ImageNet benchmark with ResNet50 model, our method achieves an average reduction of 11.15$\%$ in the false positive rate at 95$\%$ recall (FPR95) compared to the current state-of-the-art approach. The code would be released.
\end{abstract}

}

\maketitle

\IEEEdisplaynontitleabstractindextext

%
\IEEEpeerreviewmaketitle

\IEEEraisesectionheading{\section{Introduction}\label{sec:introduction}}
\IEEEPARstart{I}{n} an open-world setting, a reliable system should not only provide accurate predictions but also issue appropriate warnings when encountering
unknown data. 
Consequently, the field of out-of-distribution (OOD) detection has emerged as an essential area of research alongside the rapid development of deep neural networks (DNNs).
Typically, DNNs are assumed to be trained and tested on datasets drawn from the same distribution. This assumption allows DNNs to generate precise predictions on test data, known as in-distribution (ID) data. However, in open-world scenarios, this assumption is frequently invalidated. 
For instance, in autonomous driving, it is impractical for the training dataset to encompass all potential scenarios. Therefore, the presence of OOD inputs necessitates the ability to recognize them for a DNN to be reliable.
In this context, the development of algorithms for OOD detection holds substantial practical significance.

A rich line of research has been developed for OOD detection. For instance, output-based methods have revealed that the prediction confidence of ID data is generally higher than that of OOD data. Consequently, multiple score functions have been devised based on model outputs \cite{msp, energy, odin}. Additionally, several studies have investigated the disparities in intermediate features between ID and OOD data \cite{react,bats,maha,knn}. These studies have 
found that features of ID data tend to cluster together and differ significantly from those of OOD data. Building upon this observation, various score functions based on features, such as Maha \cite{maha} and KNN \cite{knn}, have been proposed for OOD detection.

All the aforementioned methods utilize forward information, \textit{i.e.}, information generated in the forward process, to detect OOD inputs. 
\textcolor{black}{However, the disparities between ID and OOD data are also significant in the backward process of DNNs \cite{igoe2022useful,gradnorm,lee2023probing,sun2022gradient,lee2022gradient,szolnoky2022interpretability}.}
For instance, the parameter gradient magnitude for ID data is
generally smaller than that for OOD data, given the well-trained nature of DNNs on ID data. This observation has inspired several intriguing OOD detection methods, such as GradNorm \cite{gradnorm}, which uses
the gradient norm as its score function. However, a comprehensive investigation into the divergence of parameter gradients between ID and OOD data is still lacking. The primary challenge arises from \textcolor{black}{the high dimensionality of parameter gradients in modern DNNs.}
\textcolor{black}{The distance measurement within the high dimensional space is inherently unreliable due to the "curse of dimensionality" \cite{zimek2012survey}, which refers to} the tendency for distances between data points to become increasingly similar as the number of dimensions increases, thereby rendering the distinction between ID and OOD data difficult to discern in such space.

\begin{figure*}[ht]
    \centering
    \includegraphics[width=\textwidth,height=0.52\textwidth]{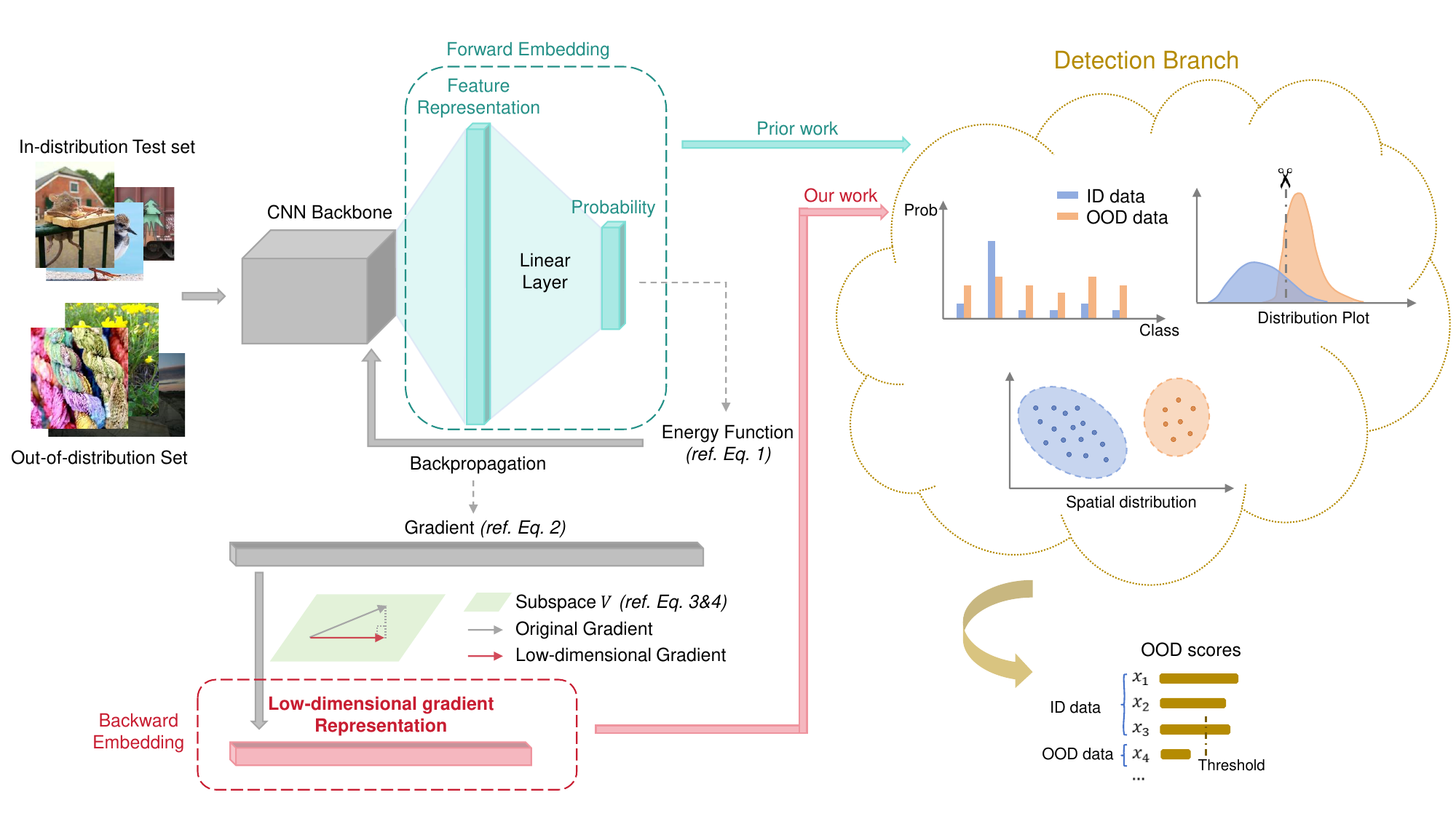} 
    \caption{\textbf{Illustration of our framework using low-dimensional gradients for OOD detection.} Firstly, we obtain the parameter gradients from a complete backpropagation process of our model. Then, using a pre-extracted subspace where the principal components of training data gradients reside, we obtain low-dimensional representations through a projection operation. Finally, we feed the representations into the detection branch, where diverse score functions are designed based on these representations.}
    \label{framework}
\end{figure*}

\begin{table*}[ht]
\vspace{-0.5em}
    \centering
    \caption{Simple modifications and superior performance when our low-dimensional gradients are integrated with six baseline methods. We report the FPR95 and AUROC metrics on the ImageNet benchmark \textcolor{black}{with ResNet50 model}. \textcolor{black}{Lower FPR95 and higher AUROC indicate better performance.} Comparison results with baseline methods are presented in brackets, with blue indicating that our method outperforms the baseline.
    \label{modification}}
    \begin{tabular*}{0.655\linewidth}{@{  }lcccc@{}}
    \toprule
        Method & Type & Simple Modification & FPR95$\downarrow$ & AUROC$\uparrow$ \\
        \midrule
        MSP \cite{msp} & Output-based & An additional linear network & 32.28 \textcolor{blue}{($\downarrow$32.38)} & 91.87 \textcolor{blue}{($\uparrow$9.05)} \\
        Energy \cite{energy} & Output-based & An additional linear network & 27.90 \textcolor{blue}{($\downarrow$29.57)} & 93.09 \textcolor{blue}{($\uparrow$6.04)}\\
        ReAct \cite{react} & Feature-based & Adaptive threshold & 23.03 \textcolor{blue}{($\downarrow$7.67)} & 95.45 \textcolor{blue}{($\uparrow$2.15)}\\
        BTAS \cite{bats} & Feature-based &  Adaptive threshold & 23.42 \textcolor{blue}{($\downarrow$34.13)} & 95.40 \textcolor{blue}{($\uparrow$8.36)}\\
        Maha \cite{maha} & Feature-based & None & 87.30 \textcolor{lightgray}{($\uparrow$5.79)} & 49.31 \textcolor{lightgray}{($\downarrow$14.78)}\\
        KNN \cite{knn} & Feature-based & None & 41.18 \textcolor{blue}{($\downarrow$15.12)} & 88.10 \textcolor{blue}{($\uparrow$2.84)} \\
    \midrule
    Ensemble & - & - & \textbf{19.55\textcolor{blue}{($\downarrow$11.15)}} & \textbf{96.12\textcolor{blue}{($\uparrow$2.82)}} \\ 
    \bottomrule
    \end{tabular*}
    \vspace{-0.6em}
\end{table*}

To address the challenge posed by high dimensionality, we propose to perform dimension reduction on the gradient. Admittedly, dimension reduction involves the potential loss of crucial information associated with significant dimensions, which may undermine its efficacy in OOD detection. Nevertheless, recent studies indicate that the gradient of ID data may primarily fall into a relatively low-dimensional space. The related theoretical discussions and empirical evaluations include:
\begin{itemize}
    \item Subspace training \cite{dldr,gur2018gradient}: During the training process, the gradient of network parameters falls into a data-dependent, low-dimensional subspace;
    \item Low-rank adaptation \cite{LoRA}: The change of network parameters for a fine-tuning task is a low-rank adaptation matrix;
    \item Low-rank spectrum of neural tangent kernel \cite{baratin2021implicit,papyan2020traces,canatar2021spectral}: The neural tangent kernel matrix, consisting of gradient inner products, has a low-rank spectrum.
\end{itemize}
These discussions shed light on an important insight: parameter gradients lie in a relatively low-dimensional space, offering the possibility of linear dimension reduction while preserving significant information for effective OOD detection. The key problem now revolves around identifying the subspace where the principal components of gradients reside. One approach we use comes from NFK \cite{nfk}, which employs an efficient algorithm to compute the top eigenvectors of a gradient covariance matrix. However, this method incurs significant computational costs due to the high dimensionality of gradients. To mitigate this burden, we 
propose an alternative method to estimate the PCA subspace. Building on the observation that gradients of samples from
the same class exhibit a pronounced degree of directional similarity \cite{papyan2020traces, fort2019emergent}, we suggest using the average gradient of each class to compose the subspace.
\textcolor{black}{This approach is easier to implement, and our empirical results in Sec. \ref{Low-dimensional Gradient Extraction} demonstrate that it can represent the top-C (C is the number of classes of ID dataset) principal components of PCA.}
In practical applications, users can flexibly choose the above two approaches based on data volume and model size \textcolor{black}{considering that PCA subspace is more accurate but requires longer computational time.} Once the subspace is determined, the gradients can be projected onto it, yielding low-dimensional representations. These representations can then be effectively utilized to distinguish between ID and OOD samples.

Given that the reduced gradients serve as a distinctive form of data representation \cite{mu2019gradients,szolnoky2022interpretability}, the approaches that use forward embeddings can be seamlessly integrated with our reduced gradients. The possible choices include MSP\cite{msp} and Energy\cite{energy}, which utilize output differences; ReAct\cite{react} and BATS\cite{bats}, which employ feature distribution differences; and Mahalanobis \cite{maha} and KNN \cite{knn}, which use pairwise distance information. With simple modifications to these approaches (see Table \ref{modification}), our reduced gradients can be effectively applied on them and show remarkable improvements in detection performance. Additionally, we discover that an ensemble of forward and backward information can further improve our detection performance since data samples exhibiting minimal disparities in the forward features may exhibit substantial variations in the backward gradients.

The overall framework of our method is illustrated in Figure \ref{framework}. To evaluate its effectiveness, we conduct experiments on two widely used benchmark datasets: CIFAR10 \cite{cifar10} and ImageNet \cite{imagenet}, along with seven OOD datasets \textcolor{black}{and six different architectures}. We evaluate the performance of the methods \cite{msp,energy,react,bats,knn,maha} mentioned above based on forward features and backward gradients respectively. The experimental results demonstrate the superior performance of our low-dimensional gradients on both benchmarks, see Table \ref{modification}. \textcolor{black}{Notably, on the large-scale ImageNet benchmark with ResNet50 model, our method outperforms the best baseline by $11.15\%$ in terms of False Positive Rate at $95\%$ True Positive Rate (FPR95)}, and by $2.82\%$ in terms of Area Under the Receiver Operating Characteristic Curve (AUROC).
The contributions of this paper can be summarized as follows:
\begin{itemize}
    \item We present a pioneering investigation into the utilization of complete gradient information for OOD detection, offering novel insights into this field.
    \item Our proposed gradient dimension reduction algorithm offers a valuable technical contribution, providing a solid foundation for future studies on leveraging gradients in OOD detection.
    \item We demonstrate the remarkable performance of our low-dimensional gradients across a diverse range of OOD detection tasks.
\end{itemize}

\section{Related Work}
The existing OOD detection approaches can be categorized into three types: 1) density-based; 2) post-hoc; 3) confidence enhancement. 

\textbf{Density-based} methods explicitly model the ID data with some probabilistic models and flag samples in low-density regions as OOD data. For example, Mahalanobis \cite{maha} uses class-conditional Gaussian distribution to model the ID data and detect OOD samples with their likelihoods. Flow-based methods \cite{kobyzev2020normalizing,zisselman2020deep,kingma2018glow,jiang2021revisiting} are also effective in probabilistic modeling for OOD detection. Besides, researchers also investigate the effect of an ensemble of multiple density models \cite{choi2018waic}. However, due to the difficulty of optimizing generative models, density-based methods are hardly used in practice.

\textbf{Post-hoc} methods are the most popular in OOD detection because of their portability and effectiveness. These approaches aim to design a score function that assigns high scores to ID data and low scores to OOD data, enabling the separation of OOD samples by applying an appropriate threshold. Based on the way of calculating scores, these methods can be broadly divided into three types: 1) output-based; 2) feature-based; and 3) gradient-based. Classical output-based methods, such as MSP\cite{msp}, ODIN\cite{odin}, and Energy\cite{energy}, design score functions based on the observation that classification probabilities of OOD samples tend to be lower than those of normal data. In pursuit of enhanced detection performance, ReAct \cite{react} and BATS \cite{bats} propose to truncate the abnormal values of intermediate features, thereby amplifying the disparities between ID and OOD data in the output space. Feature-based approaches, as exemplified by KNN \cite{knn}, Maha \cite{maha}, and other related works \cite{ndiour2020out,cook2020outlier}, primarily leverage the aggregation property of ID features to identify OOD samples. For instance, Maha \cite{maha} utilizes the Mahalanobis distance between the feature of inputs and the mean feature of corresponding classes as its score function for OOD detection. Similarly, KNN \cite{knn} employs the Euclidean distance to the $k$-th nearest neighborhood as a measurement to detect OOD data. Furthermore, several studies \cite{ndiour2020out,cook2020outlier} leverage the inherent low-dimensional characteristics of features and introduce reconstruction errors as detection scores to identify OOD data. Gradient-based approaches, such as GradNorm \cite{gradnorm}, Purview \cite{lee2023probing}, GraN \cite{lust2020gran}, and others \cite{lee2022gradient,sun2022gradient,igoe2022useful}, mainly focus on utilizing the norm of parameter gradients to detect OOD samples. For instance, GradNorm \cite{gradnorm} shows that the gradient norm calculated from the categorical cross-entropy loss is generally lower on OOD data, thereby designing detection functions based on the norm. Other studies employ the gradient norm in different ways, like introducing binary networks \cite{lee2022gradient}, etc \cite{lee2023probing,lust2020gran,sun2022gradient,igoe2022useful}. Apart from the above-mentioned approaches, Vim \cite{wang2022vim} empirically demonstrates that a combination of output and feature information can further improve detection performance. Our proposed method also belongs to the post-hoc type, which is plug-and-play in practice. 

\textbf{Confidence enhancement} methods design various regularization terms to amplify the difference between ID and OOD data. 
\textcolor{black}{One type of them is to utilize auxiliary OOD datasets to fine-tune the model with designed training losses that explicitly enlarge the output difference between ID and auxiliary OOD data. The most classical method is outlier exposure \cite{oe}, which encourages the output of OOD data to be uniformly distributed.}
Another type is to apply contrastive losses \textcolor{black}{on ID dataset} to promote stronger ID-OOD separability, like SimCLR \cite{chen2020simple}, SupCon \cite{khosla2020supervised}, and Cider\cite{ming2022cider}. However, confidence enhancement methods typically require model retraining, and in some cases, even necessitate the acquisition of additional OOD data. These requirements render such methods less practical for real-world applications.

\section{Method\label{method}}
\subsection{Preliminary}
The framework of OOD detection can be described as follows. We consider a classification problem with $C$ classes, where $\mathcal{X}$ stands for the input space and $\mathcal{Y}$ for the label space. The joint data distribution over $\mathcal{X} \times \mathcal{Y}$ is denoted as $D_{\mathcal{X}\mathcal{Y}}$. Let $f_\theta$ : $\mathcal{X}\mapsto \mathcal{Y}$ be a model trained on samples drawn \textit{i.i.d.} from $D_{\mathcal{X}\mathcal{Y}}$ with parameter $\theta$. Then the distribution of ID data is denoted as $D_{in}$, which is the marginal distribution of $D_{\mathcal{X}\mathcal{Y}}$ over $\mathcal{X}$. The distribution of OOD data is presented as $D_{out}$, whose label set has no intersection with $\mathcal{Y}$. The goal of OOD detection is to decide whether a test input $x$ is from $D_{in}$ or $D_{out}$. For post hoc methods, the decision is made through a score function $S$ as follows:
\begin{equation*}
G_\lambda(x) = \left\{
    \begin{aligned}
    &\text{ID} &\quad& \text{if}\ S(x,f) \geq \lambda, \\
    &\text{OOD} &\quad& \text{if}\ S(x,f)\textless \lambda, \\
    \end{aligned}
\right.
\end{equation*}
where $\lambda$ is a threshold and samples with scores higher than $\lambda$ are classified as ID data. The threshold is usually set based on ID data to guarantee that a high fraction of ID data (\textit{e.g.} 95$\%$) is correctly identified as ID samples.

\subsection{Low-dimensional Gradient Extraction}\label{Low-dimensional Gradient Extraction}
In this section, we progressively answer two questions: 1) how to calculate gradients in the label-agnostic condition? 2) how to reduce the gradient dimension and simultaneously retain as much information as possible? In the context of OOD detection, input labels are inaccessible to users, and thus the normal cross-entropy loss used in the training process is not applicable to calculating gradients for test data. Borrowing the idea from JEM \cite{grathwohl2019your}, we induce a label-free energy function that has the same conditional probability $p_\theta(y|x)=\exp(f^y_\theta(x))/\sum_y\exp(f^y_\theta(x))$ as the original model:
\begin{equation}
    E(x;\theta)=-\log\sum_y\exp(f^y_\theta(x))
\end{equation}
It essentially reframes a conditional distribution over $y$ given $x$ to an induced unconditional distribution over $x$. With the energy-based model $E(x;\theta)$, we can calculate the gradient given any input $x$ without $y$ as follows:
\begin{equation}
    \nabla_\theta E(x;\theta) = -\sum_y p_\theta(y|x)\nabla_\theta f_\theta^y(x)
\end{equation}
It is equivalent to a weighted average of the derivative of outputs for all labels. To eliminate the scale discrepancies among different dimensions, we normalize it using the mean and variance matrix calculated from the training data:
\begin{equation*}
    \begin{aligned}
    & \mathcal{M} = \mathbb{E}_{x\sim D_{in}}\frac{\partial E(x;\theta)}{\partial \theta},\quad \mathcal{I}=\mathbb{E}_{x\sim D_{in}}[U(x)U(x)^T]\\
    & G(x)=(diag(\mathcal{I})^{-\frac{1}{2}})(\nabla_\theta E(x;\theta) - \mathcal{M})\\
    \end{aligned}
\end{equation*}
where $\mathcal{M}$ and $\mathcal{I}$ are the mean and variance matrix, and $G(x)$ is the normalized gradient of sample $x$. After obtaining the label-agnostic gradient, the next crucial problem is reducing its dimension to facilitate its practical utilization. Recent empirical studies, such as LoRA \cite{LoRA} and DLDR \cite{dldr}, have shown that the gradient of ID data resides in a low-dimensional subspace. The low-rank spectrum analysis \cite{baratin2021implicit, papyan2020traces, canatar2021spectral} of neural tangent kernel (NTK, \cite{ntk}) has also provided theoretical evidence explaining the above phenomenon. Therefore, a simple linear dimension reduction method is sufficient for extracting a low-dimensional representation of the gradient. The most commonly used algorithm is principal component analysis (PCA, \cite{wold1987principal}), which employs the top-K eigenvectors of the gradient covariance matrix as its dimension reduction matrix. Denote the eigendecomposition as follows:
\begin{equation}\label{pca}
    \begin{aligned}
        & C = V\Sigma V^T,
    \end{aligned}
\end{equation}
where $C$ is the gradient covariance matrix computed on all the training data. Based on the diagonal values of $\Sigma$, PCA selects $K$ column vectors of $V$ corresponding to the first $K$ large eigenvalues. Then the low-dimensional representation is calculated as follows:
\begin{equation*}
    \begin{aligned}
        & G_{low} = GV.
    \end{aligned}
\end{equation*}
Because of the high dimensionality of gradients, the above eigen decomposition cannot be calculated directly in practice. Hence, we adopt a power iteration method \cite{bathe1971solution,golub2000eigenvalue} to obtain the eigenvectors under reasonable computation costs, following the idea in NFK \cite{nfk}. Specific algorithms are in Appendix A. 

In addition to the PCA algorithm, in this paper, we also propose an alternative approach to estimate the low-dimensional subspace encompassing principal components. Drawing upon the observation that the parameter update induced by a sample has a more pronounced influence on samples belonging to the same class while exerting minimal influence on samples from other classes \cite{charpiat2019input,he2019local,fort2019stiffness}, it is reasonable to infer that parameter gradients from samples within the same class tend to align in similar directions. Previous empirical studies \cite{papyan2020traces,fort2019emergent} support the above inference by demonstrating that parameter gradients from samples of the same class exhibit a high degree of direction similarity. Moreover, our empirical experiments, detailed in Appendix B, further validate the above claim by revealing a high cosine similarity between the gradients of samples from the same class and their corresponding gradient centers. Based on these findings, we propose utilizing the average gradient (AG) of each class as an alternative subspace estimation:
\begin{equation}\label{grad center}
    V\triangleq\{\mathbb{E}_{(x,y)\sim D_{\mathcal{X}\mathcal{Y}}}[G(x| y=c)]\},\ c=1,2,\ldots,C.
\end{equation}
In this way, the extraction of low-dimensional subspace becomes easy to implement. However, this subspace is inaccurate compared to the PCA one. \textcolor{black}{Specifically, we measure the reconstruction error of gradients of ID training and test data to the PCA and AG subspace, which reflects the amount of information lost of gradients after projecting them into the low-dimensional subspace. The results in Table \ref{2-1} indicate that the gradient information is partially lost in the AG subspace while almost completely retained in the PCA subspace. Furthermore, the similar error of AG subspace and 10-dimensional PCA subspace implies that the average gradient direction is consistent with the top-C (C is the number of classes of ID dataset) principal components of PCA, which demonstrates the rationality of using AG subspace to reduce the gradient dimension. Considering the computational cost of PCA, the AG subspace is more suitable for scenarios involving large volumes of data and complex models.
}
\begin{table}[ht]\color{black}
    \centering
    \caption{Reconstruction error of gradients for Train and Test data to different subspaces. The error is defined as $\Vert G-GVV^T\Vert_2/\Vert G \Vert_2$. We report the mean and variance of reconstruction error for 1000 randomly selected samples on each dataset.}
    \label{2-1}
    \begin{tabular}{c|c|c|c}
    \toprule
        Subspace & Dim & Train & Test \\
        \midrule
        Avg Grad & 10 & 0.6569$_{\pm 0.0163}$ & 0.6791$_{\pm 0.0237}$ \\
        PCA & 10 & 0.6004$_{\pm 0.0141}$ & 0.6165$_{\pm 0.0225}$ \\
        PCA & 200 & 0.0790$_{\pm 0.0116}$ & 0.0841$_{\pm 0.0214}$ \\
    \bottomrule
    \end{tabular}
    \vspace{-0.5em}
\end{table}

Based on the above two subspace extraction methodologies, the overall algorithm for calculating the low-dimensional gradient is shown in Alg \ref{reduced gradient}. It is worth noticing that although the subspace extraction process takes some computing time, the follow-up projection process can be fast with efficient parallel calculation. Specifically, the subspace can be partitioned into multiple lower-dimensional subspaces, each of which can be stored on different GPUs. Then, the projection process can be independently executed on each GPU. By subsequently concatenating the outcomes obtained from each GPU, the final low-dimensional gradient can be obtained efficiently. \textcolor{black}{Specific analysis about the computation time can be seen in Sec \ref{Time Overhead}.}

\begin{algorithm}[ht]
\caption{Low-dimensional Gradient Calculation}\label{reduced gradient}
\begin{algorithmic}[1]
\REQUIRE model $f_\theta$, sample $x$, mean $\mathcal{M}$, variance $\mathcal{I}$
\ENSURE The low-dimensional gradient of $x$
\STATE Extract the principal subspace $V$ using Eq (\ref{pca}) or Eq (\ref{grad center}).
\STATE $G_{org} \gets (\text{diag}(\mathcal{I})^{-\frac{1}{2}})(-\sum_y p_\theta(y|x)\nabla_\theta f_\theta^y(x)-\mathcal{M})$
\STATE $G_{low} \gets G_{org}V$
\STATE \textbf{return}  $G_{low}$
\end{algorithmic}
\end{algorithm}

\subsection{Score Function Design}
\begin{figure}[ht]
    \centering
    \subfloat{\includegraphics[width=0.2\textwidth,height=0.15\textwidth]{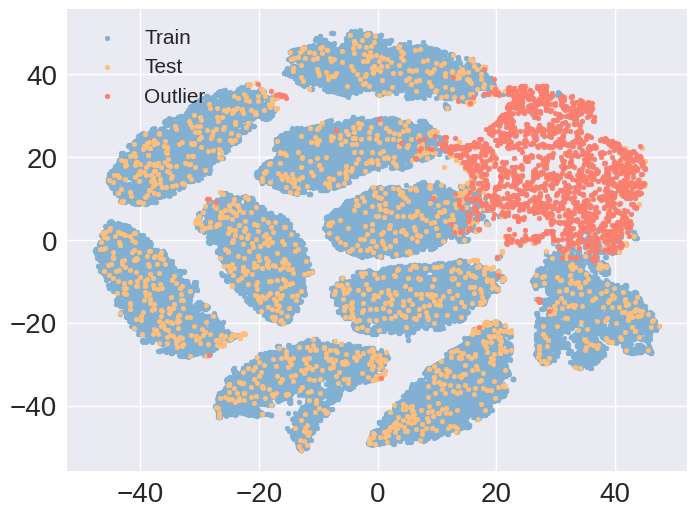}}
    \hspace{0.5cm}
    \subfloat{\includegraphics[width=0.2\textwidth,height=0.15\textwidth]{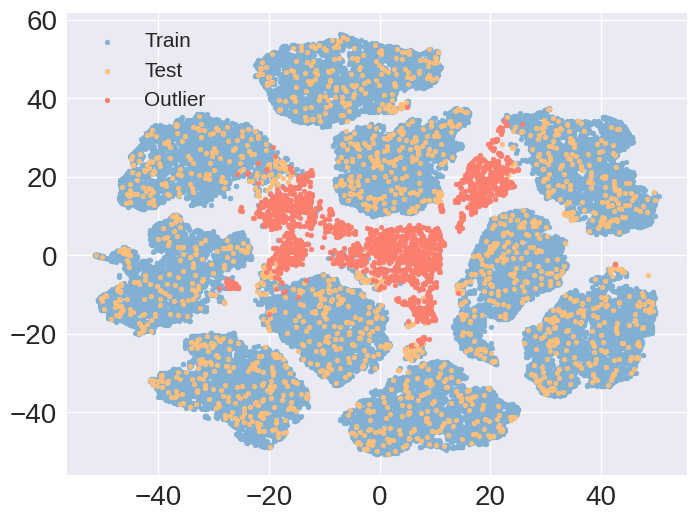}}
    \caption{Visualization of penultimate features (left) and low-dimensional gradients (right) via t-SNE \cite{van2008visualizing}. The model is trained with ResNet18 on CIFAR10. \label{visualization}}
\end{figure}

\begin{figure}[ht]
  \centering
  \subfloat{\includegraphics[width=0.2\textwidth,height=0.15\textwidth]{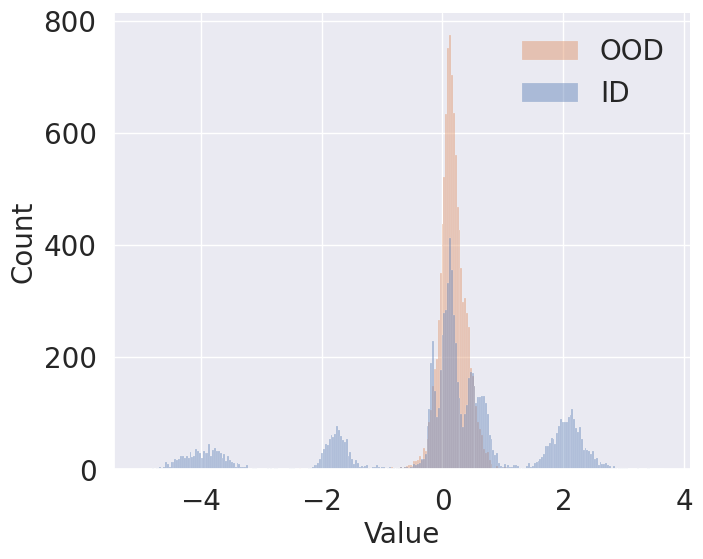}\label{1}}
  \hspace{0.5cm}
  \subfloat{\includegraphics[width=0.2\textwidth,height=0.15\textwidth]{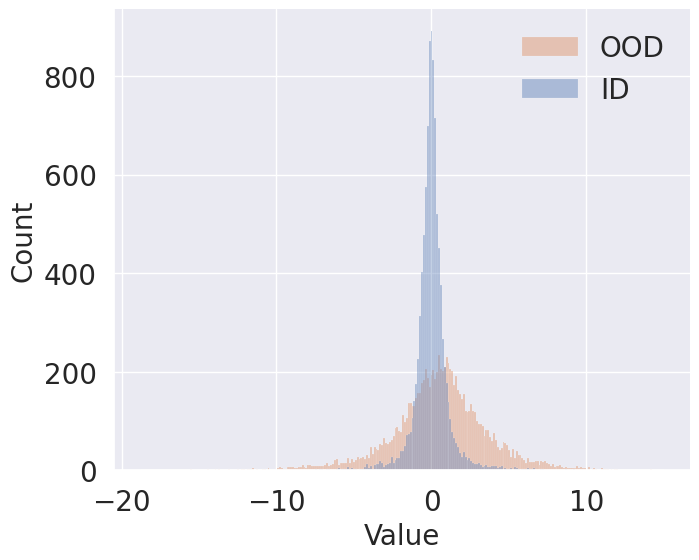}\label{2}}
  \vspace{0.3cm}
  \subfloat{\includegraphics[width=0.2\textwidth,height=0.15\textwidth]{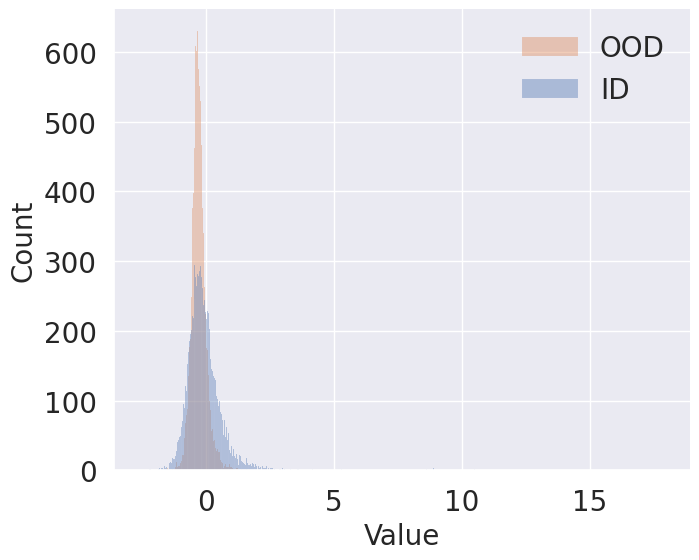}\label{3}}
  \hspace{0.5cm}
  \subfloat{\includegraphics[width=0.2\textwidth,height=0.15\textwidth]{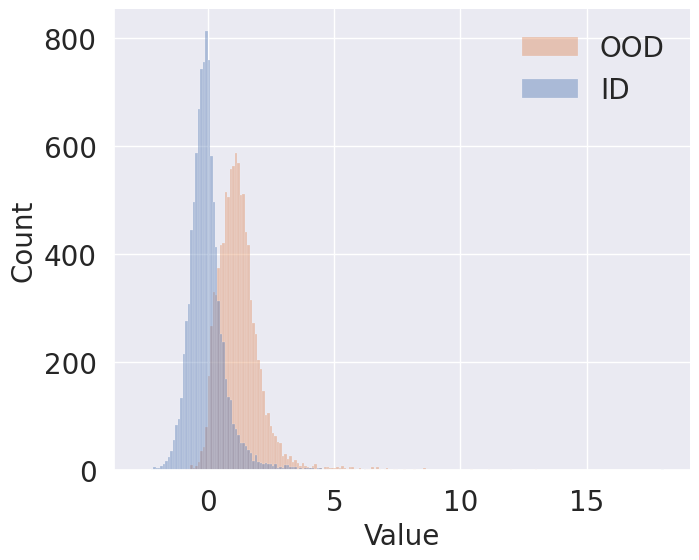}\label{4}}
  \caption{Density distribution of low-dimensional gradients on CIFAR10 and ImageNet on different dimensions. Top left: The 9th dimension of our reduced gradient on CIFAR10. Top right: The 11th dimension on CIFAR10. Bottom left: The 1th dimension on ImageNet. Bottom right: The 999th dimension on ImageNet. These plots reveal that the distribution differences between ID and OOD data is more distinguishable on the later dimensions.}
  \label{gradient distribution}
\end{figure}

This section elucidates the application of low-dimensional gradients in OOD detection. To design effective score functions based on the gradients, we first analyze their properties to figure out the divergence between ID and OOD data. The visualization of our low-dimensional gradients presented in Figure \ref{visualization} reveals a discernible phenomenon: the low-dimensional gradients derived from ID data tend to exhibit clustering, forming cohesive groups, while simultaneously manifesting a distinctive separation from the gradients originating from OOD data. This observation demonstrates a striking parallel to the clustering and separation patterns observed in forward features. Therefore, prior distance-based methods such as Maha \cite{maha} and KNN \cite{knn} can be seamlessly adapted to operate on our gradients. Furthermore, since our gradients show the separability between samples of different classes, they can serve as a distinctive form of data representation and be used to classify samples. Thus, output-based methods \cite{msp, energy} can also be combined with our gradients. Apart from the visualization analysis, we also investigate the distribution of our gradients to dig up their distinctions between ID and OOD data. As Figure \ref{gradient distribution} shows, the values of OOD gradients fall within the range of ID gradients in the top-n dimensions (n is a number close to the number of classes $C$), while in the rest dimensions, their values are noticeably larger. Hence, we can employ previous feature rectification methods \cite{react,bats} on the rest dimensions of our gradients to obtain improved detection performance compared to the output-based methods.

Subsequently, we provide detailed explanations regarding the integration of our gradients with output-based methods \cite{msp,energy}, feature rectification methods \cite{react,bats}, and distance-based methods \cite{maha,knn}. Besides, inspired by the ensemble strategy proposed in Vim \cite{vim}, we also explore a simple ensemble technique that combines forward and backward information to achieve better OOD detection performance.

\subsubsection{Combination with output-based methods}
Output-based approaches aim to capture the dissimilarities in feature representations between ID and OOD data by modulating the network output using a linear layer. Consequently, we introduce an auxiliary linear network, trained on our reduced gradient, to generate the necessary output for computing score functions. The network architecture is:\\
\begin{equation}\label{linear network}
    g\rightarrow \text{BN} \rightarrow \text{FC} \rightarrow y
\end{equation}
\textcolor{black}{The input is a low-dimensional gradient and the output is a $C$-dimensional tensor that indicates the prediction probability of each class ($C$ is the number of categories of ID data). To train the network, we adopt a common training setting, where low-dimensional gradients of ID training data as the training dataset, the cross-entropy loss with the label of training data as the loss function, and the stochastic gradient descent (SGD) as the optimization approach. 
}
The simplicity of the network enables a rapid training process. Once the linear network has been effectively trained, we can leverage its output to discriminate between ID and OOD samples. The basic score function is the maximum softmax confidence \cite{msp}, \textit{i.e.},
\begin{equation}\label{s1}
    S_1(x,f) = \max_j \frac{e^{y_j}}{\sum_j^C e^{y_j}}.
\end{equation}
Furthermore, the energy \cite{energy} score function is proved to be more efficient in OOD detection, which quantifies the free energy function associated with each sample:
\begin{equation}\label{s2}
    S_2(x;f)=-T\cdot \log\sum_j^C e^{y_j/T}
\end{equation}
Both $S_1$ and $S_2$ exhibit large values on ID data and vice versa.

\subsubsection{Combination with feature rectification methods}
To further amplify the output difference between ID and OOD data, prior studies have explored the distribution of penultimate features and proposed feature rectification techniques, as OOD data is observed to exhibit larger feature values \cite{react, bats}. However, our experimental findings indicate that applying these methods directly to our reduced gradients yields inferior performance compared to output-based approaches. Based on the distribution observation in Figure \ref{gradient distribution}, we propose to exclusively rectify our low-dimensional gradients in certain dimensions and then feed them into the aforementioned linear network in (\ref{linear network}). The overall process can be expressed as:
\begin{equation}
    g \rightarrow \text{BN} \rightarrow \text{Clip(}g_d\text{)} \rightarrow \text{FC} \rightarrow y
\end{equation}
where $g\in R^{K}$ and we denote $g_d\in R^{d}$ as the last-$d$ dimensional vector of $g$. Different rectification approaches employ different Clip functions to truncate the values of gradients. For example, ReAct \cite{react} method truncates activations above a threshold $c$ to limit the effect of noises:
\begin{equation*}
    \text{Clip(}g_d\text{)}=\min(g_d,c)
\end{equation*}
where $c$ a threshold that is determined based on the $p$-th percentile of $g_d$ estimated on ID data. For example, when $p=90$, it indicates that 90$\%$ percent of ID gradients are less than $c$. Apart from ReAct approach, BATS \cite{bats} method proposes to rectify extreme activations to the boundary values of typical sets with the guidance of batch normalization:
\begin{equation*}
    \text{Clip(}g_d\text{)}= 
    \begin{cases}
      \mu+\lambda\delta & \text{if  } g_d-\mu \geq \lambda\delta \\
      \text{BN}(g_d) & \text{if  } -\lambda\delta < g_d-\mu < \lambda\delta \\
       \mu-\lambda\delta & \text{if  } g_d-\mu \leq -\lambda\delta
    \end{cases}\\
\end{equation*}
where $\mu$ and $\delta$ are the learnable bias and weight of batch normalization layer, respectively, and $\lambda$ is a hyperparameter. The typical set is defined as the interval [$\mu-\lambda\delta$, $\mu+\lambda\delta$], within which all activations are constrained to reside. After the rectification of gradients, both the ReAct and BATS methods utilize the energy score function (Eq. (\ref{s2})) to detect OOD samples.

\subsubsection{Combination with distance-based methods}
Features of ID data tend to cluster together and be away from those of OOD data; thus, previous works designed various distance measurements as score functions to detect OOD data. For our low-dimensional gradients, these methods should be equally effective since gradients share the same aggregation and separation properties as features, as shown in Figure \ref{visualization}. One typical measurement is the Mahalanobis distance \cite{maha} as follows:
\begin{equation}\label{s3}
    S_3(x,f)= \max_c -(g-\hat{\mu_c})^T\hat{\Sigma}^{-1}(g-\hat{\mu_c})
\end{equation}
where $\hat{\mu_c}$ and $\hat{\Sigma}$ are the empirical class mean and covariance of training samples. The Mahalanobis distance-based method imposes a class-conditional Gaussian distribution assumption about the underlying gradient space, while another distance-based approach named KNN\cite{knn} is more flexible and general without any distributional assumptions. It utilizes the Euclidean distance to the $k$-th nearest neighbor of training data as its score function, which can be expressed as follows:
\begin{equation}\label{s4}
    S_4(x,f)= -\Vert g-\hat{g}_{(k)} \Vert_2
\end{equation}
Both $S_3$ and $S_4$ functions exhibit larger values on ID data and smaller values on OOD data since they take the negative value of the distance measurements.

\subsubsection{Ensemble of forward and backward information}\label{method ensemble}
Drawing inspiration from the concept of integrating multiple sources of information to enhance detection performance, as demonstrated in \cite{vim}, our work also tries to combine the scores obtained from the feature/output space with those derived from our reduced gradients. The algorithm for this combination is mathematically expressed as follows:
\begin{equation}\label{ensemble}
    S_5(x,f) = S_{f}(x,f) + \alpha * S_{b}(x,f)
\end{equation}
where $\alpha$ is the weight coefficient to ensure that $S_f$ and $\alpha*S_b$ are of similar magnitude. In this paper, we set $\alpha$ as 1 in our experiments.

\section{Experiments}
In this section, we conduct a comparative analysis of the performance attained by employing gradients and features independently across six distinct detection methods. The analysis is performed on two widely recognized benchmark datasets, namely CIFAR10 \cite{cifar10} and ImageNet \cite{imagenet}, \textcolor{black}{with six different network architectures}. Additionally, we evaluate another three advanced detection approaches to validate our superior performance in OOD detection. Moreover, we assess the detection performance of utilizing the combined information from both forward and backward sources, which further reduces FPR95 and improves AUROC. At the end of this section, we investigate the impact of reduced dimensionality on our detection performance as well as the influence of various hyperparameters in the underlying methods. Through a comprehensive series of experiments, we provide compelling evidence for the effectiveness of employing low-dimensional gradients in OOD detection, thereby offering novel insights for future research endeavors. The detailed experimental configurations and settings are presented below.\\
\hspace*{\fill} \\
\noindent
\textbf{OOD Datasets}
We evaluate our method on two popular benchmarks: CIFAR10 \cite{cifar10} and ImageNet \cite{imagenet}. When using CIFAR10 as ID data, we consider four OOD datasets commonly used in literature \cite{knn,ming2022exploit,react,dal,li2023hierarchical,msp,energy}: Textures \cite{texture}, SVHN \cite{svhn}, LSUN-C \cite{lsunc}, and iSUN \cite{isun}. As for the evaluation on ImageNet, we choose four commonly used OOD datasets that are subsets of: iNaturalist \cite{inat}, SUN \cite{sun}, Places \cite{places}, and Texture \cite{texture} with non-overlapping categories \textit{w.r.t.} ImageNet.\\
\hspace*{\fill} \\
\noindent
\textbf{Evaluation Metrics}
Two classical metrics are reported in this paper: 1) FPR95: the false positive rate of OOD samples when the true positive rate of ID samples is at 95$\%$. 2) AUROC: the area under the receiver operating characteristic curve. The smaller the FPR95 and the higher the AUROC, the better the performance.\\
\hspace*{\fill} \\
\noindent
\textbf{Models and Hyper-parameters}
In the case of CIFAR10, \textcolor{black}{we respectively adopt the ResNet18 \cite{he2016deep}, MobileNet-v2 \cite{sandler2018mobilenetv2}, DenseNet121 \cite{huang2017densely} and Wideresnet-28-10 \cite{zagoruyko2016wide} as our base model. }
We train them using the SGD algorithm with weight decay 0.0005, momentum 0.9, cosine schedule, initial learning rate 0.1, epoch 200, and batch size 128. \textcolor{black}{For MobileNet-v2 architecture, we change its stride from 2 to 1 to make it more suitable for CIFAR10 dataset. Noticing the high ratio of infinite values in the variance matrix $\mathcal{I}$ for MobileNet-v2 and DenseNet121, we excluded $\mathcal{I}$ from gradient normalization for these models.} The pivotal hyper-parameter of our low-dimensional gradient is the reduced dimensionality, denoted as $K$. \textcolor{black}{For the ResNet18, MobileNet-v2 and DenseNet architectures, we adopt the PCA subspace with $K=200$ to reduce the gradient dimension. For the Wideresnet-28-10 model, we adopt the average gradient subspace considering that it has much more parameters.}
Regrading the ImageNet dataset, \textcolor{black}{we use the pre-trained model in Pytorch \cite{pytorch} with different architectures including ResNet50 \cite{he2016deep}, MobileNet-v2 \cite{sandler2018mobilenetv2}, DenseNet121 \cite{huang2017densely} and ViT-B16 \cite{dosovitskiy2020image}. The top-1 test accuracy of them are respectively $76.13\%$, $71.88\%$, $74.43\%$ and $85.42\%$. We adopt the average gradient subspace for all the models to obtain the low-dimensional gradient.}\\
\hspace*{\fill} \\
\noindent
\textbf{Baseline Methods}
In this paper, we apply gradients and features separately within six different detection methods, including two output-based methods, namely MSP \cite{msp} and Energy \cite{energy}, two feature rectification methods, namely ReAct \cite{react} and BATS \cite{bats}, and two distance-based methods, namely Mahanobias \cite{maha} and KNN \cite{knn}. Additionally, we validate the superiority of our proposed approach by comparing it with three other advanced detection methods, namely ODIN \cite{odin}, GradNorm \cite{gradnorm}, and Vim \cite{vim}. Notably, all of the aforementioned approaches employ pre-trained networks in a post hoc manner. 

\subsection{Evaluation on Large-scale ImageNet Benchmark}
We first evaluate our method on the large-scale ImageNet benchmark, which poses significant challenges due to the substantial volume of training data and model parameters. The comparison results are presented in Table \ref{imagenet}, where we evaluate six different detection methods \textcolor{black}{on four networks} using either features or our low-dimensional gradients as inputs. 
\textcolor{black}{Given the poor performance of the Mahalanobis method \cite{maha} on the ResNet50 model, we do not evaluate it on other networks. Additionally, since ViT uses layer normalization instead of batch normalization, we exclude the BATS method \cite{bats} from testing on the ViT model.} To control our computational overhead, we randomly select 50000 training samples as the training dataset for our detection algorithms. We train the linear network in Eq.(\ref{linear network}) with $0.01$ learning rate, $512$ batch size, and $3$ epochs, achieving a test accuracy of $74.24\%$ on ResNet50, \textcolor{black}{$71.07\%$ on MobileNet-v2, $73.98\%$ on DenseNet121 and $82.78\%$ on ViT.} For ReAct and BATS, we choose to truncate the last 50 dimensions of the 1000-dimensional gradients with $p=0.7$ and $\lambda=0.1$. Regarding the KNN method, we set $K=10$ for both forward and backward embeddings.

It is worth noticing that \textcolor{black}{our reduced gradient integrated with the simplest detection algorithm MSP always achieves remarkable performance.} For example, on the ResNet50 architecture, our method has FPR95 of $32.38\%$ and AUROC of $91.87\%$, reducing the FPR95 by $32.38\%$ and improving the AUROC by $9.05\%$ compared to the original MSP. Moreover, combing with feature rectification methods like ReAct \cite{react} and BATS \cite{bats}, our performance can be further improved. For instance, on ResNet50 model, our approach achieves SOTA \textcolor{black}{average} performance with FPR95 of $23.03\%$ and AUROC of $95.45\%$ based on ReAct method, surpassing the best baseline by $7.67\%$ in FPR95 and $2.15\%$ in AUROC. \textcolor{black}{One expectation is for SUN dataset, where ReAct works better for features, indicating that the choice of embedding is indeed task-dependent. But overall our low-dimensional gradients are quite promising for distinguishing ID and OOD data.}

\begin{figure}[!t]
    \centering
    \subfloat{\includegraphics[width=0.2\textwidth,height=0.15\textwidth]{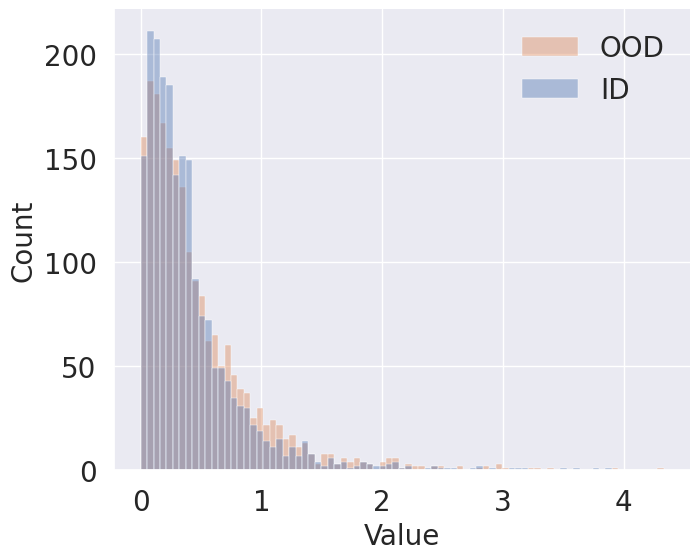}}
    \hspace{0.5cm}
    \subfloat{\includegraphics[width=0.2\textwidth,height=0.15\textwidth]{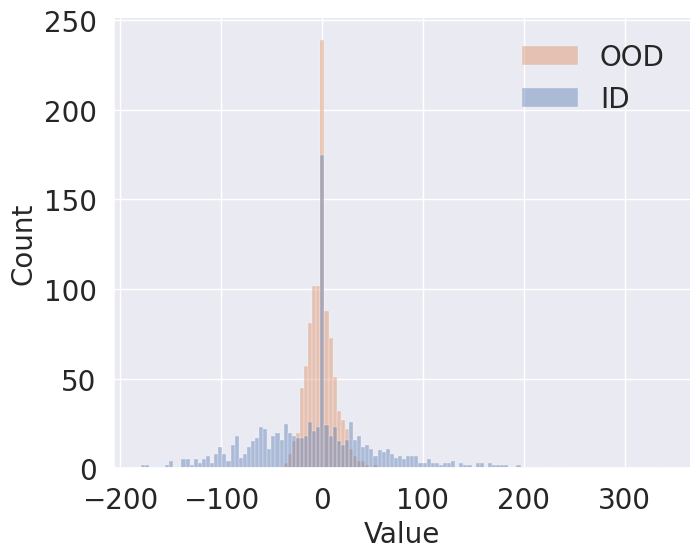}}
    \caption{Density distribution of features (left) and low-dimensional gradients (right) of two samples (one is OOD data from Places and the other is ID data from ImageNet). We can observe that the distinction between the two samples is marginal in the feature space but significant in the gradient space.}
    \label{distribution_places}
\end{figure}

There is an impeccable performance of our method on the Places OOD dataset with ResNet50 model, which implies a significant disparity in the gradient space between the ImageNet and Places datasets. To gain further insight into this disparity, we randomly choose an OOD sample in the Places dataset and an ID sample in the ImageNet dataset and analyze the density distribution of their forward features and reduced gradients. The result presented in Figure \ref{distribution_places} demonstrates that their distinctions are marginal in the feature space but pronounced in the gradient space, which aligns with the performance we achieved on the Places dataset. 

\begin{table*}[!t]\color{black}
  \centering
  \caption{Results on ImageNet benchmark with various network architectures. Six different methods are evaluated respectively using forward features and backward gradients. The best result is in bold.\label{imagenet}}
  \begin{tabular*}{\linewidth}{@{  }cccccccccccc@{}}
    \toprule
    \multirow{2}{*}{Methods} & \multirow{2}{*}{Embedding} & \multicolumn{2}{c}{iNaturalist} & \multicolumn{2}{c}{SUN} & \multicolumn{2}{c}{Places} & \multicolumn{2}{c}{Texture} & \multicolumn{2}{c}{Average} \\
      & & FPR95$\downarrow$ & AUROC$\uparrow$ & FPR95$\downarrow$ & AUROC$\uparrow$ & FPR95$\downarrow$ & AUROC$\uparrow$ & FPR95$\downarrow$ & AUROC$\uparrow$ & FPR95$\downarrow$ & AUROC$\uparrow$ \\
     \midrule
     \multicolumn{12}{c}{ResNet50, ID Acc $76.13\%$} \\
     \midrule
     \multirow{2}{*}{MSP} & Feature & 52.77 & 88.42 & 68.58 & 81.75 & 71.57 & 80.63 & 66.13 & 80.46 & 64.76 & 82.82 \\
     & Gradient & 42.20 & 90.05 & 67.82 & 81.45 & 0.00 & 99.99 & 19.50 & 95.98 & 32.38 & 91.87\\
     \midrule
     
     \multirow{2}{*}{Energy} & Feature & 53.93 & 90.59 & 58.27 & 86.73 & 65.40 & 84.13 & 52.29 & 86.73 & 57.47 & 87.05 \\
     & Gradient & 28.79 & 93.39 & 67.52 & 82.26 & 0.00 & 100.00 & 15.28 & 96.70 & 27.90 & 93.09 \\
     \midrule
     
     \multirow{2}{*}{ReAct} & Feature & 19.49 & 96.40 & 24.07 & 94.40 & 33.48 & 91.92 & 45.74 & 90.47 & 30.70 & 93.30 \\
     & Gradient & 19.87 & 95.77 & 45.34 & 90.52 & 0.00 & 100.00 & 26.90 & 95.50 & \textbf{23.03} & \textbf{95.45} \\
     \midrule
     
     \multirow{2}{*}{BATS} & Feature & 54.12 & 90.59 & 58.31 & 86.74 & 65.39 & 84.14 & 52.38 & 86.70 & 57.55 & 87.04\\
     & Gradient & 20.23 & 95.69 & 45.40 & 90.53 & 0.00 & 100.00 & 28.05 & 95.38 & 23.42 & 95.40 \\
     \midrule
     
     \multirow{2}{*}{Maha} & Feature & 93.70 & 65.81 & 97.56 & 49.91 & 97.63 & 49.44 & 37.13 & 91.18 & 81.51 & 64.09\\
     & Gradient & 95.41 & 47.93 & 96.88 & 52.66 & 100.00 & 37.87 & 56.90 & 58.78 & 87.30 & 49.31 \\
     \midrule
     
     \multirow{2}{*}{KNN} & Feature & 63.85 & 85.40 & 70.46 & 81.63 & 76.27 & 77.52 & 14.63 & 96.49 & 56.30 & 85.26 \\
     & Gradient & 42.76 & 91.20 & 86.98 & 66.92 & 33.06 & 94.94 & 10.90 & 97.30 & 43.43 & 87.59 \\
    \midrule
    \multicolumn{12}{c}{MobileNet-v2, ID Acc $71.88\%$} \\ 
    \midrule
    \multirow{2}{*}{MSP} & Feature &  59.83 & 86.72 & 74.18 & 78.88 & 76.88 & 78.14 & 70.99 & 78.95 & 70.47 & 80.67 \\
    & Gradient & 42.94 & 91.27 & 61.20 & 85.72 & 67.94 & 82.84 & 61.45 & 80.77 & 58.38 & 85.15 \\ 
    \midrule
    \multirow{2}{*}{Energy} & Feature & 55.31 & 90.34 & 59.36 & 86.24 & 66.27 & 83.21 & 54.52 & 86.58 & 58.87 & 86.59 \\
    & Gradient & 39.65 & 92.81 & 60.77 & 86.71 & 66.66 & 83.93 & 53.12 & 86.45 & 55.05 & 87.48 \\
    \midrule
    \multirow{2}{*}{ReAct} & Feature & 45.16 & 92.40 & 53.24 & 87.58 & 60.95 & 84.39 & 41.08 & 90.85 & 50.11 & 88.81 \\
    & Gradient & 34.80 & 93.22 & 50.09 & 89.44 & 57.05 & 87.02 & 40.19 & 90.91 & 45.53 & 90.15 \\
    \midrule
    \multirow{2}{*}{BATS} & Feature & 55.31 & 90.34 & 59.36 & 86.24 & 66.27 & 83.21 & 54.52 & 86.58 & 58.87 & 86.59\\
    & Gradient & 32.42 & 93.35 & 46.96 & 89.72 & 53.65 & 87.30 & 39.60 & 90.93 &  \textbf{43.16} &  \textbf{90.33} \\
    \midrule
    \multirow{2}{*}{KNN} & Feature & 99.81 & 29.23 & 98.90 & 46.94 & 99.42 & 42.40 & 47.85 & 85.47 & 86.50 & 51.01 \\
    & Gradient & 50.68 & 89.44 & 74.61 & 76.74 & 78.88 & 73.59 & 57.09 & 88.94 & 65.32 & 82.18 \\ 
    \midrule
    \multicolumn{12}{c}{DenseNet121, ID Acc $74.43\%$} \\
    \midrule
    \multirow{2}{*}{MSP} & Feature & 49.26 & 89.05 & 67.04 & 81.54 & 69.22 & 81.05 & 67.07 & 79.19 & 63.15 & 82.71 \\
     & Gradient & 38.20 & 89.96 & 67.90 & 78.91 & 50.54 & 86.17 & 54.60 & 82.75 & 52.81 & 84.45 \\
     \midrule
     \multirow{2}{*}{Energy} & Feature & 39.69 & 92.66 & 51.98 & 87.40 & 57.84 & 85.17 & 52.11 & 85.42 & 50.41 & 87.66 \\
     & Gradient & 34.50 & 91.34 & 67.70 & 79.96 & 48.12 & 88.40 & 51.90 & 85.73 & 50.56 & 86.36 \\
     \midrule
     \multirow{2}{*}{ReAct} & Feature & 28.08 & 94.51 & 45.34 & 90.27 & 52.70 & 87.18 & 47.75 & 89.31 & 43.47 & 90.32 \\
     & Gradient & 24.40 & 95.46 & 58.50 & 85.56 & 39.03 & 91.97 & 40.71 & 91.88 & 40.66 & 91.22 \\
     \midrule
     \multirow{2}{*}{BATS} & Feature & 39.69 & 92.66 & 51.98 & 87.40 & 57.84 & 85.17 & 52.11 & 85.42 & 50.41 & 87.66 \\
     & Gradient & 22.14 & 95.78 & 54.30 & 85.69 & 35.29 & 92.08 & 40.10 & 92.01 &  \textbf{37.96} &  \textbf{91.39} \\
     \midrule
     \multirow{2}{*}{KNN} & Feature & 99.95 & 26.91 & 98.99 & 42.09 & 99.38 & 39.14 & 57.13 & 80.13 & 88.86 & 47.07 \\
     & Gradient & 58.39 & 90.20 & 97.30 & 36.94 & 70.56 & 83.31 & 53.76 & 85.94 & 70.00 & 74.10 \\ %
    \midrule
    \multicolumn{12}{c}{ViT-base-p16-384, ID Acc $85.42\%$} \\
    \midrule
    \multirow{2}{*}{MSP} & Feature & 19.04 & 96.11 & 56.74 & 86.10 & 60.08 & 85.04 & 48.55 & 87.10 & 46.10 & 88.59 \\
     & Gradient & 13.80 & 96.96 & 56.60 & 84.10 & 50.60 & 90.61 & 43.10 & 92.34 & 41.03 & 91.00 \\
    \midrule
    \multirow{2}{*}{Energy} & Feature & 6.16 & 98.66 & 36.93 & 91.68 & 45.38 & 89.18 & 28.22 & 93.39 & 29.17 & 93.23 \\
    & Gradient & 5.80 & 98.42 & 35.82 & 91.79 & 43.76 & 91.31 & 22.14 & 94.99 & 26.88 & 94.13 \\
    \midrule
    \multirow{2}{*}{ReAct} & Feature & 5.58 & 98.63 & 43.36 & 90.33 & 50.99 & 87.85 & 33.23 & 90.98 & 33.29 & 91.95 \\
    & Gradient & 5.40 & 98.56 & 31.40 & 95.63 & 37.00 & 94.05 & 21.20 & 94.48 &  \textbf{23.75} & \textbf{95.68} \\
    \midrule
    \multirow{2}{*}{KNN} & Feature & 48.70 & 91.33 & 77.50 & 77.61 & 72.50 & 78.87 & 43.00 & 88.64 & 60.43 & 84.11 \\
    & Gradient & 40.30 & 91.53 & 74.40 & 77.36 & 68.40 & 80.25 & 41.91 & 89.02 & 56.25 & 84.54 \\
  \bottomrule
  \end{tabular*}
\end{table*}

\subsection{Evaluation on CIFAR10 Benchmark}
To further demonstrate the effectiveness of our low-dimensional gradients, we also evaluate our method on the CIFAR10 benchmark. The comparison results are presented in Table \ref{cifar10}. The hyper-parameters of baseline methods are set to the recommended values, i.e., $T=1$ for Energy, $p=0.9$ for ReAct, $\lambda=0.1$ for BATS, and $K=5$ for KNN. The Mahalanobis method we tested solely relies on the original distance formulation in Eq.(\ref{s3}) without incorporating any input pre-processing techniques or feature ensemble approaches. Thus, its performance is expected to be inferior to what has been reported in previous works where such enhancements were employed. Comparing all the results, our proposed gradients integrated with the KNN \cite{knn} method demonstrate superior performance \textcolor{black}{on diverse models}, \textit{e.g.}, achieving the smallest FPR95 of $21.55\%$ and the highest AUROC of $96.33\%$ on ResNet18. \textcolor{black}{Notably, the Mahalanobis method shows a terrible performance on the MobileNet-v2 model, which implies its high sensitivity to network architectures. In contrast, the performance of our method is consistently outstanding across different models.}

\begin{table}[!t]\color{black}
  \centering
  \caption{Results on CIFAR10 benchmark with various network architectures. We report the averaged FPR95 and AUROC across selected OOD datasets. The best result is in bold. \label{cifar10}}
  \resizebox{\linewidth}{!}{
  \begin{tabular}{cccccccc}
    \toprule
    Methods & MSP & Energy & ReAct & BATS & Maha & KNN & Ours \\
    \midrule
    \multicolumn{8}{c}{ResNet18, ID Acc $96.39\%$}\\
    \midrule
    FPR95$\downarrow$ & 33.08 & 29.47 & 36.46 & 21.66 & 22.27 & 27.59 & \textbf{21.55} \\
    AUROC$\uparrow$ & 90.06 & 89.40 & 83.34 & 96.21 & 96.31 & 95.84 & \textbf{96.33} \\
    \midrule
    \multicolumn{8}{c}{MobileNet-v2, ID Acc $92.71\%$} \\
    \midrule
    FPR95$\downarrow$ & 69.49 & 48.26 & 49.29 & 48.26 & 95.40 &  56.42 & \textbf{36.83}\\
    AUROC$\uparrow$ & 87.51 & 91.00 & 91.58 & 91.00 & 40.92 & 88.95 & \textbf{92.81} \\
    \midrule
    \multicolumn{8}{c}{DenseNet121, ID Acc $91.31\%$} \\
    \midrule
    FPR95$\downarrow$ & 63.88 & 55.33 & 66.81 & 55.33 & 54.69 & 55.80 & \textbf{53.90} \\
    AUROC$\uparrow$ & 82.13 & 77.83 & 78.08 & 77.83 & 90.11 & 90.06 & \textbf{90.61} \\
    \midrule
    \multicolumn{8}{c}{Wideresnet-28-10, ID Acc $97.03\%$} \\
    \midrule
    FPR95$\downarrow$ & 32.48 & 33.69 & 40.21 & 33.69 & 23.34 & 23.00 & \textbf{22.63}\\
    AUROC$\uparrow$ & 81.87 & 79.12 & 84.08 & 79.12 & 96.30 & 96.13 & \textbf{96.74}\\
  \bottomrule
\end{tabular}}
\end{table}

\begin{table}[!t]\color{black}
    \centering
    \caption{Results on hard detection settings including \textit{CIFAR10 v.s. CIFAR100} and \textit{CIFAR10 v.s. Tiny-ImageNet} with ResNet18 model. The best result is in bold.}
    \label{cifar10-vs-cifar100}
    \resizebox{\linewidth}{!}{
    \begin{tabular}{cccccccc}
    \toprule
        Methods & MSP & Energy & ReAct & BATS & Maha & KNN & Ours \\
        \midrule
        \multicolumn{8}{c}{CIFAR100 as OOD data and CIFAR10 as ID data} \\
        \midrule
        FPR95$\downarrow$ & 50.42 & 47.89 & 54.45 & 43.34 & 58.19 & 47.96 &  \textbf{41.28} \\
        AUROC$\uparrow$ & 81.94 & 80.87 & 72.94 & 91.15 & 84.97 & 90.45 &  \textbf{91.54} \\ 
        \midrule
        \multicolumn{8}{c}{Tiny-ImageNet as OOD data and CIFAR10 as ID data} \\
        \midrule
        FPR95$\downarrow$ & 48.03 &	44.90 & 50.78 & 41.97 & 56.10 & 44.23 &  \textbf{40.45}\\
        AUROC$\uparrow$ & 82.82 & 81.93 & 75.73 & 90.47 & 86.07 & 90.66 &  \textbf{90.92}\\
    \bottomrule
    \end{tabular}}
\end{table}

\vspace{-4.0ex}
\textcolor{black}{\subsection{Hard OOD detection}
In addition to the commonly-used OOD datasets, we further conduct experiments under the near-OOD setting proposed in \cite{yang2021generalized}. Specifically, in the setting of CIFAR10 as ID dataset, we evaluate our method on two hard OOD datasets including CIFAR100 \cite{krizhevsky2009cifar} and Tiny-ImageNet \cite{krizhevsky2017imagenet} with ResNet18 model. We adopt the reduced dimension $K=200$ and KNN basic score function \cite{knn}, the same as our common settings on CIFAR10 benchmark. The results are shown in Table \ref{cifar10-vs-cifar100}, where our method still achieves surpassing performance compared to other approaches. Specifically, our method exceeds the best baseline in terms of FPR95 by $2.06\%$ on CIFAR100 dataset and $1.52\%$ on Tiny-ImageNet dataset.
}

\subsection{Comparison with Other Methods}
In addition to the aforementioned approaches, there are other detection algorithms, such as ODIN \cite{odin}, GradNorm \cite{gradnorm}, and Vim \cite{vim}, that demonstrate outstanding performance but are not suitable for integration with our low-dimensional gradients. In this section, we compare our method with these algorithms to demonstrate our superior performance in OOD detection.

ODIN is an output-based method that incorporates the gradient of inputs as a data enhancement mechanism to obtain good detection performance. However, since that our low-dimensional gradients are not normalized to a scale of 0-1, it is difficult to determine an appropriate enhancement hyper-parameter. Hence, we do not combine ODIN with our reduced gradients.

GradNorm leverages the norm of parameter gradients as its scoring function to effectively identify OOD samples, exhibiting outstanding performance on the ImageNet benchmark dataset. Nevertheless, in our specific setting, where gradients are used as inputs to our linear network, GradNorm is not suitable.

Vim, a novel approach, capitalizes on the low dimensionality of forward features and introduces a fusion strategy that combines the reconstruction error and energy score function as its detection function. However, the high dimensionality of gradients makes it computationally intensive to calculate the reconstruction error, thereby rendering this method inappropriate to apply to the gradients. Instead, we draw inspiration from the fusion concept and propose an ensemble strategy to detect OOD samples in Section \ref{Ensemble of forward and backward information}.

\begin{figure}
    \centering
    \subfloat{\includegraphics[width=0.2\textwidth,height=0.15\textwidth]{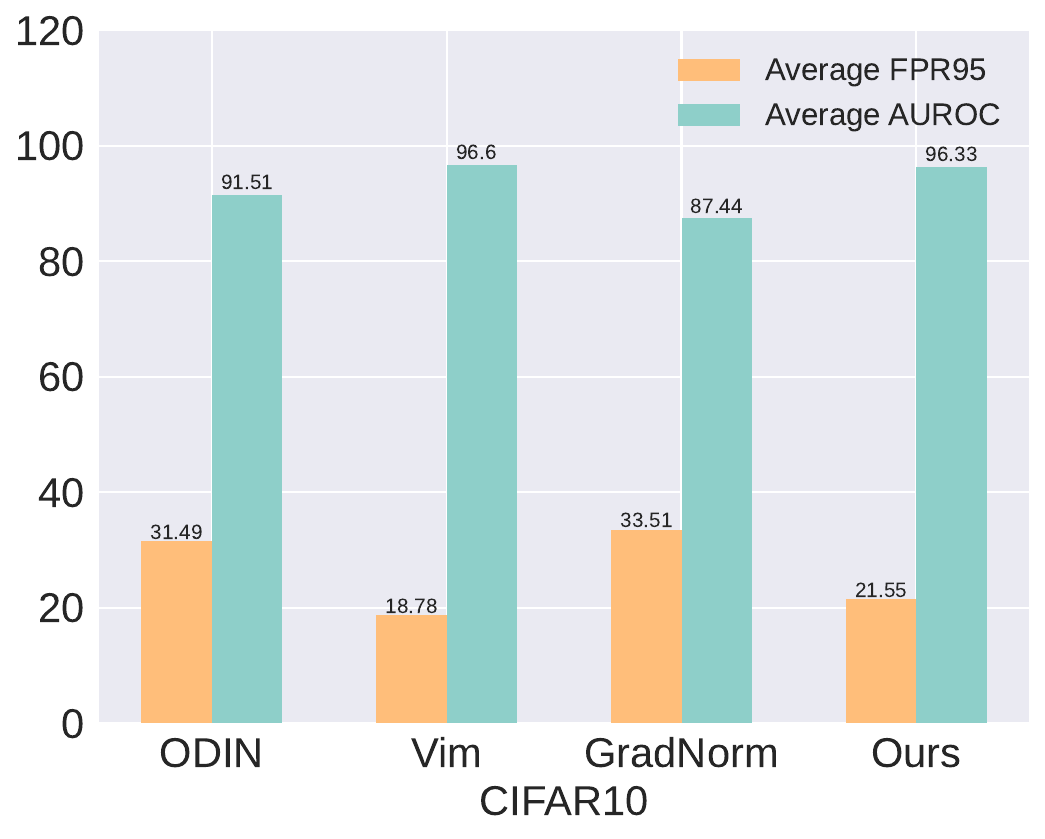}}
    \hspace{0.5cm}
    \subfloat{\includegraphics[width=0.2\textwidth,height=0.15\textwidth]{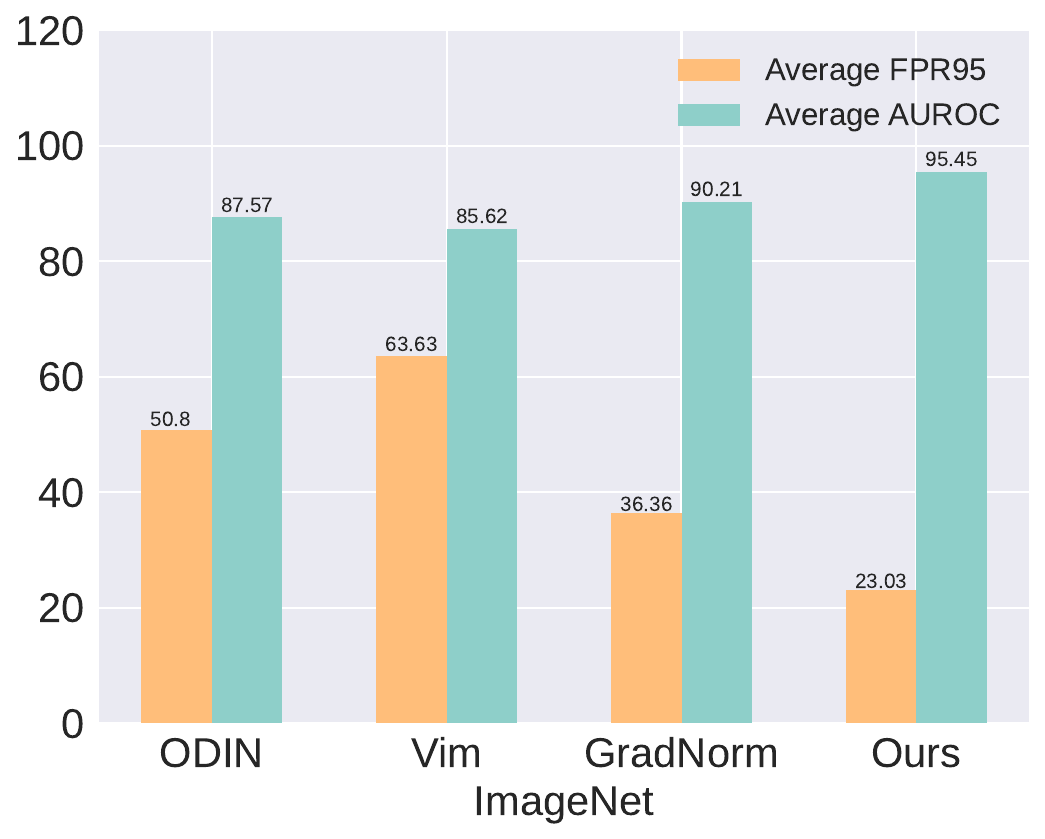}}
    \caption{Comparison with ODIN \cite{odin}, GradNorm \cite{gradnorm} and Vim \cite{vim} methods on CIFAR10 with ResNet18 model (left) and ImageNet with ResNet50 model (right).}
    \label{Comparison with Other Methods}
\end{figure}

The comparison results are presented in Figure \ref{Comparison with Other Methods}, where we use ResNet18 and ResNet50 respectively on CIFAR10 and ImageNet. Our method demonstrates comparable performance on CIFAR10 benchmark and superior performance on ImageNet benchmark, surpassing the best baseline by $5.24\%$ in AUROC.




\subsection{Ensemble of forward and backward information}\label{Ensemble of forward and backward information}
To further improve our detection performance, we explore strategies for integrating forward and backward information, as depicted in Eq.(\ref{ensemble}). We select the best-performing baseline method using our reduced gradients to conduct this ensemble approach. The experimental results are presented in Table \ref{result of ensemble}, revealing the effectiveness of information ensemble on both CIFAR10 and ImageNet benchmarks. Particularly noteworthy is the further improvement in our performance on the ImageNet dataset, achieving an FPR95 of $19.55\%$ and an AUROC of $96.12\%$.

\begin{table}[htbp]
    \centering
    \caption{Detection performance with the ensemble of forward and backward information on CIFAR10 with ResNet18 model and on ImageNet with ResNet50 model. \textcolor{black}{Values in blue brackets indicate improvement over the baseline using forward features.}\label{result of ensemble}}
    \resizebox{\linewidth}{!}{
    \begin{tabular}{cccccc}
    \toprule
        \multirow{2}{*}{Dataset} & \multirow{2}{*}{Base Method} & \multirow{2}{*}{Metric} & \multicolumn{3}{c}{Embeddings} \\
         & & & Feature & Gradient & Ensemble \\
        \midrule
        \multirow{2}{*}{CIFAR10} & \multirow{2}{*}{KNN} & FPR95$\downarrow$ & 27.59 & 21.55 & \textbf{20.24} \textcolor{blue}{($\downarrow$7.35)} \\
         & & AUROC$\uparrow$ & 95.84 & 96.33 & \textbf{96.79} \textcolor{blue}{($\uparrow$0.95)} \\
        \midrule
        \multirow{2}{*}{ImageNet} & \multirow{2}{*}{ReAct} & FPR95$\downarrow$ & 30.70 & 23.03 & \textbf{19.55} \textcolor{blue}{($\downarrow$11.15)} \\
         & & AUROC$\uparrow$ & 93.30 & 95.45 & \textbf{96.12} \textcolor{blue}{($\uparrow$2.82)} \\
    \bottomrule
    \end{tabular}}
\end{table}

\subsection{Hyper-parameter study}
In this section, we provide further analysis of the influence of hyper-parameters on our detection performance to demonstrate the stability of our approach. Specifically, we study the influence of the reduced dimension $K$ on our detection performance \textcolor{black}{and accordingly propose the strategy of choosing $K$.}
Besides, since our reduced gradients are applied in different methods, we also investigate how the hyper-parameters in these methods impact our detection performance. 

\begin{figure}[!t]
  \centering
  \subfloat{\includegraphics[width=0.2\textwidth,height=0.15\textwidth]{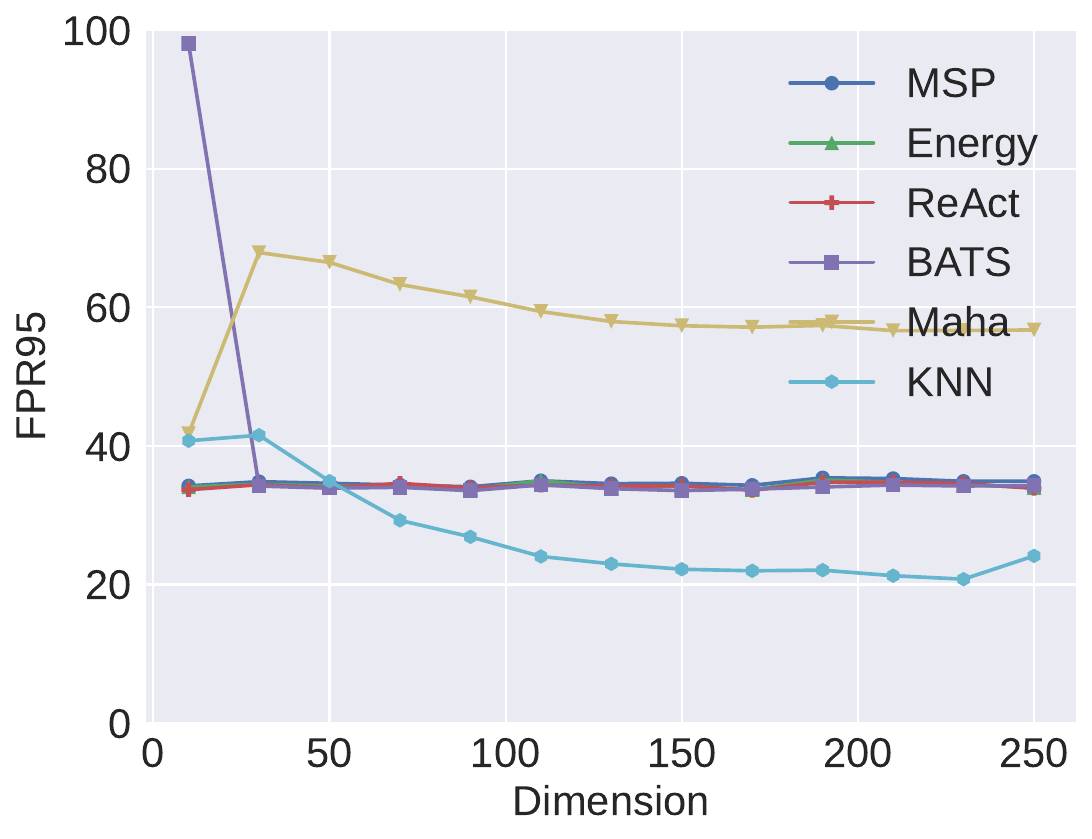}}
  \hspace{0.5cm}
  \subfloat{\includegraphics[width=0.2\textwidth,height=0.15\textwidth]{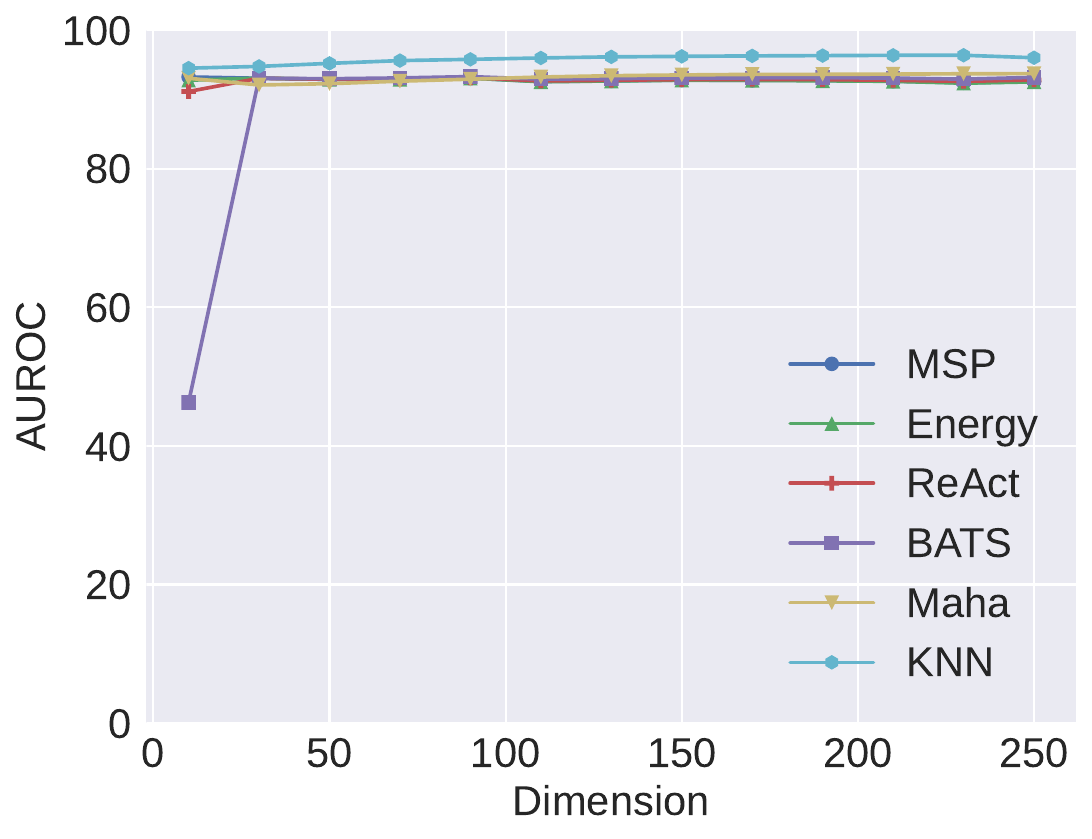}}
  \vspace{0.3cm}
  \subfloat{\includegraphics[width=0.2\textwidth,height=0.15\textwidth]{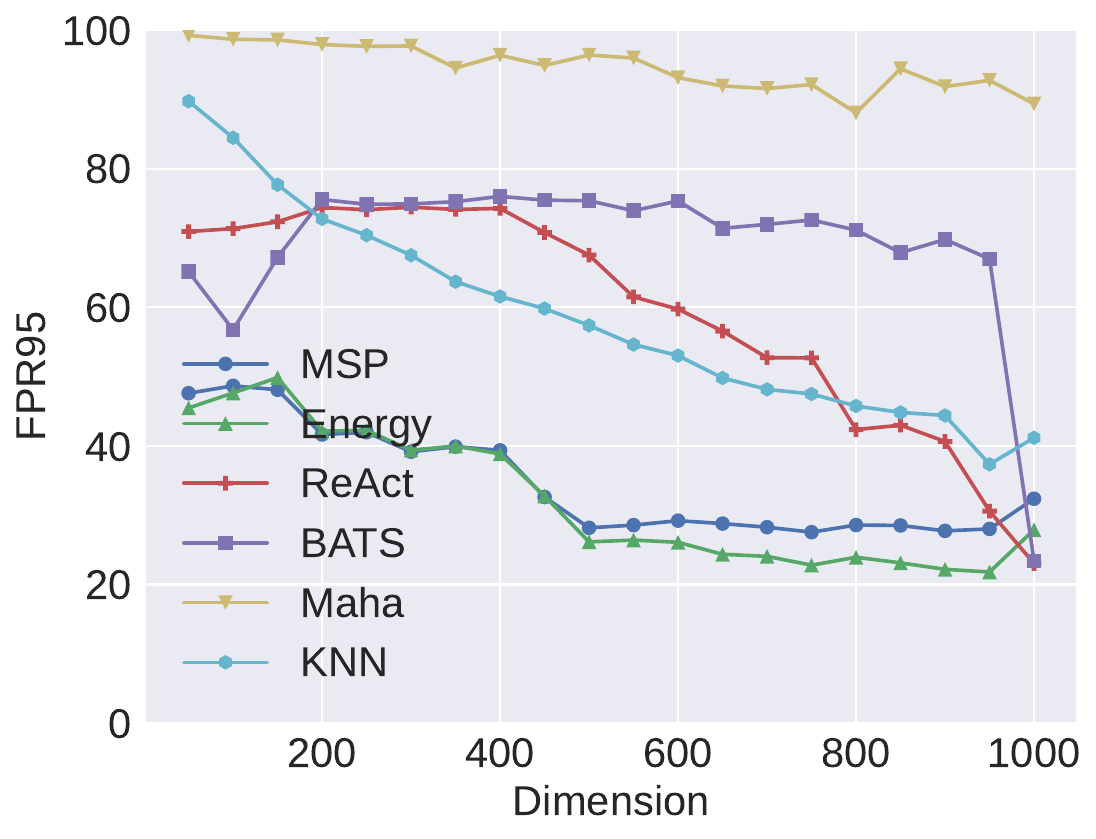}}
  \hspace{0.5cm}
  \subfloat{\includegraphics[width=0.2\textwidth,height=0.15\textwidth]{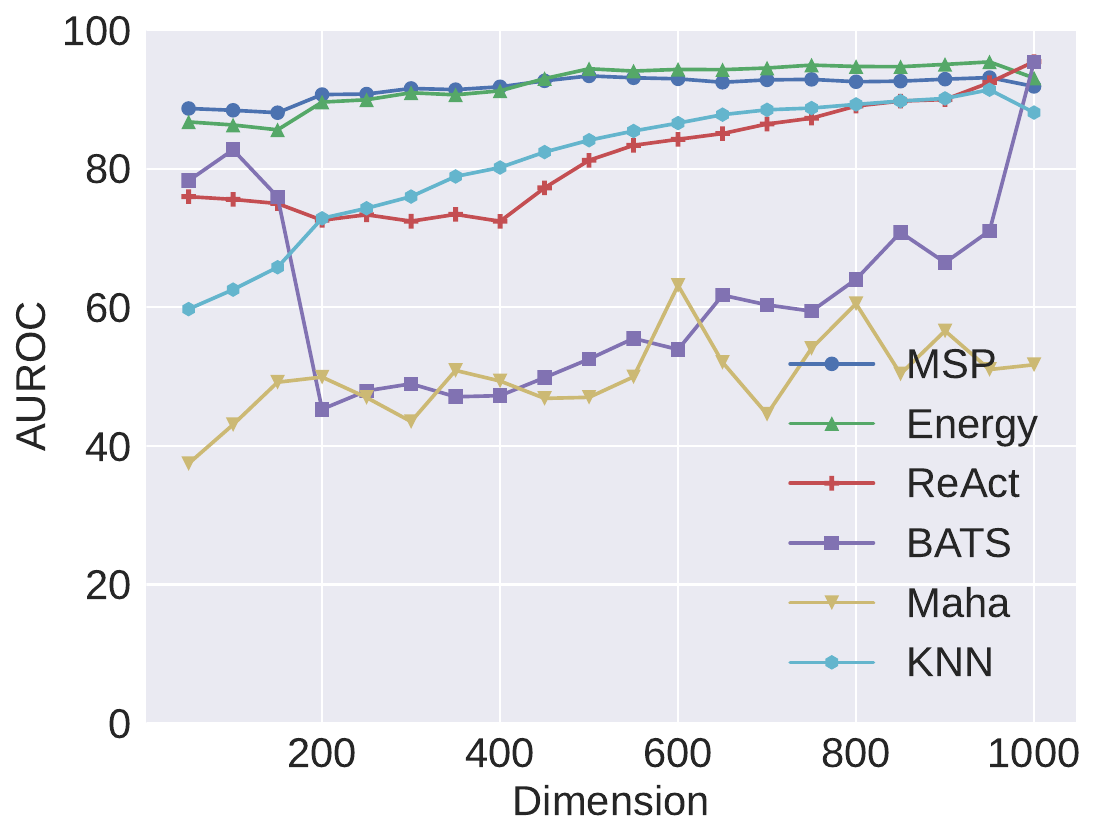}}
  \caption{Influence of  reduced dimension $K$ on our detection performance across six baseline methods on CIFAR10 (top) and ImageNet (bottom).}
  \vspace{-3ex}
  \label{effect_of_k}
\end{figure}

\textbf{Effect of K.}
In Figure \ref{effect_of_k}, we systematically analyze the effect of $K$ on six baseline methods. We vary the number of $K=\{$10, 30, 50, 70, 90, 110, 130, 150, 170, 190, 210, 230, 250$\}$ for CIFAR10 and $K=\{$50, 100, 150, 200, 250, 300, 350, 400, 450, 500, 550, 600, 650, 700, 750, 800, 850, 900, 950, 1000$\}$ for ImageNet. Our analysis reveals that the performance of our gradient-based methods improves as the dimension $K$ increases, both for the CIFAR10 and ImageNet datasets, \textcolor{black}{except that} the FPR95 of the KNN method reaches its lowest value when $K=230$ and exhibits a slight increase when $K=250$ on CIFAR10 benchmark. 

\textcolor{black}{\textbf{Choice of $K$.}}
\textcolor{black}{
The exception result of KNN score on $K=250$ is actually consistent with our ratio analysis presented in Appendix A, where we observe that the first 230 principal components account for nearly $100\%$ of the total variance. If the dimension is lower than $230$, the information of ID gradients cannot be covered. If the dimension is higher than $230$, the information on residual dimensions may be unexpectedly included, which accounts for a large proportion of OOD gradients while negligible in ID gradients. Since the difference between ID and OOD gradients can be more significant if magnitudes of OOD gradients are filtered out while magnitudes of ID gradients are retained, we do not expect to induce larger values for OOD gradients as the dimension increases. Therefore, the optimal reduction dimension should be the least dimension that can account for the total variance of ID gradients. Our experiments on CIFAR10 benchmark demonstrate that $200$ is a suitable dimension that is widely applicable on different architectures.
}

\begin{figure}[!t]
  \centering
  \subfloat{\includegraphics[width=0.2\textwidth,height=0.15\textwidth]{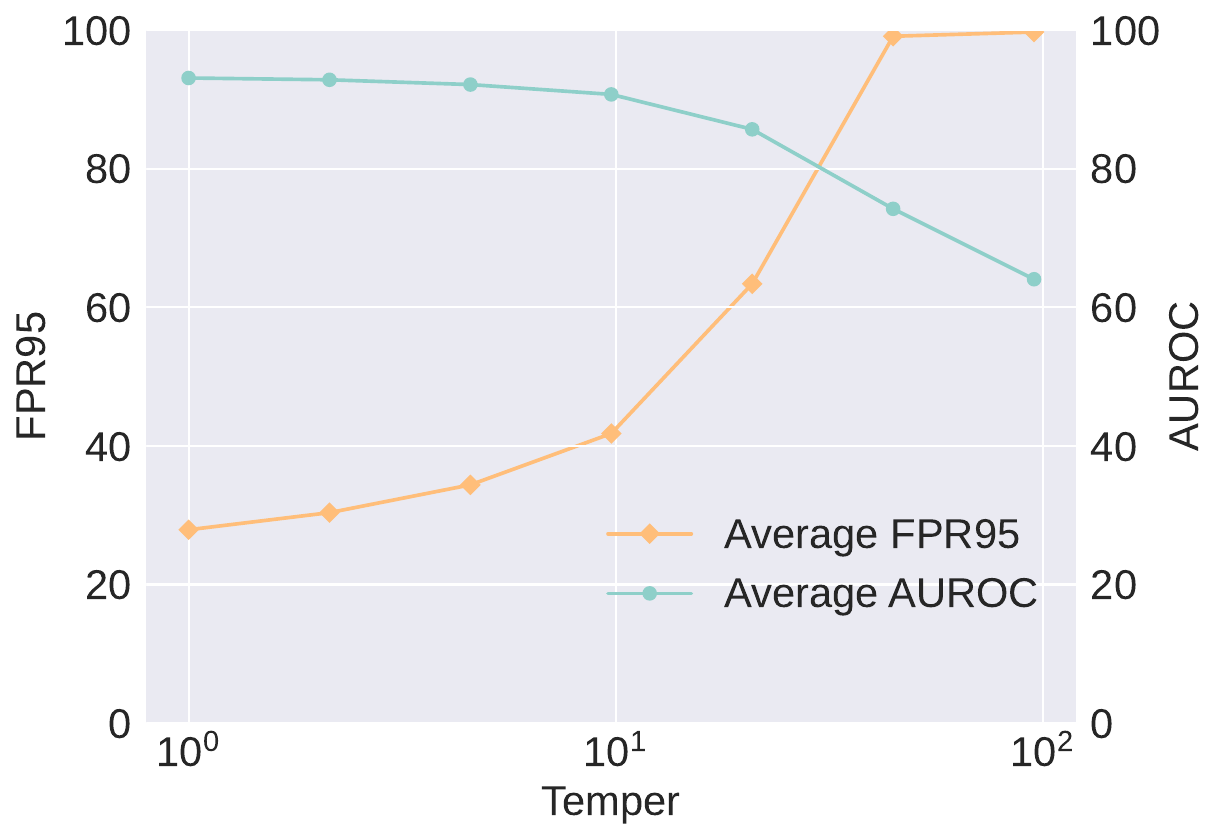}\label{1}}
  \hspace{0.5cm}
  \subfloat{\includegraphics[width=0.2\textwidth,height=0.15\textwidth]{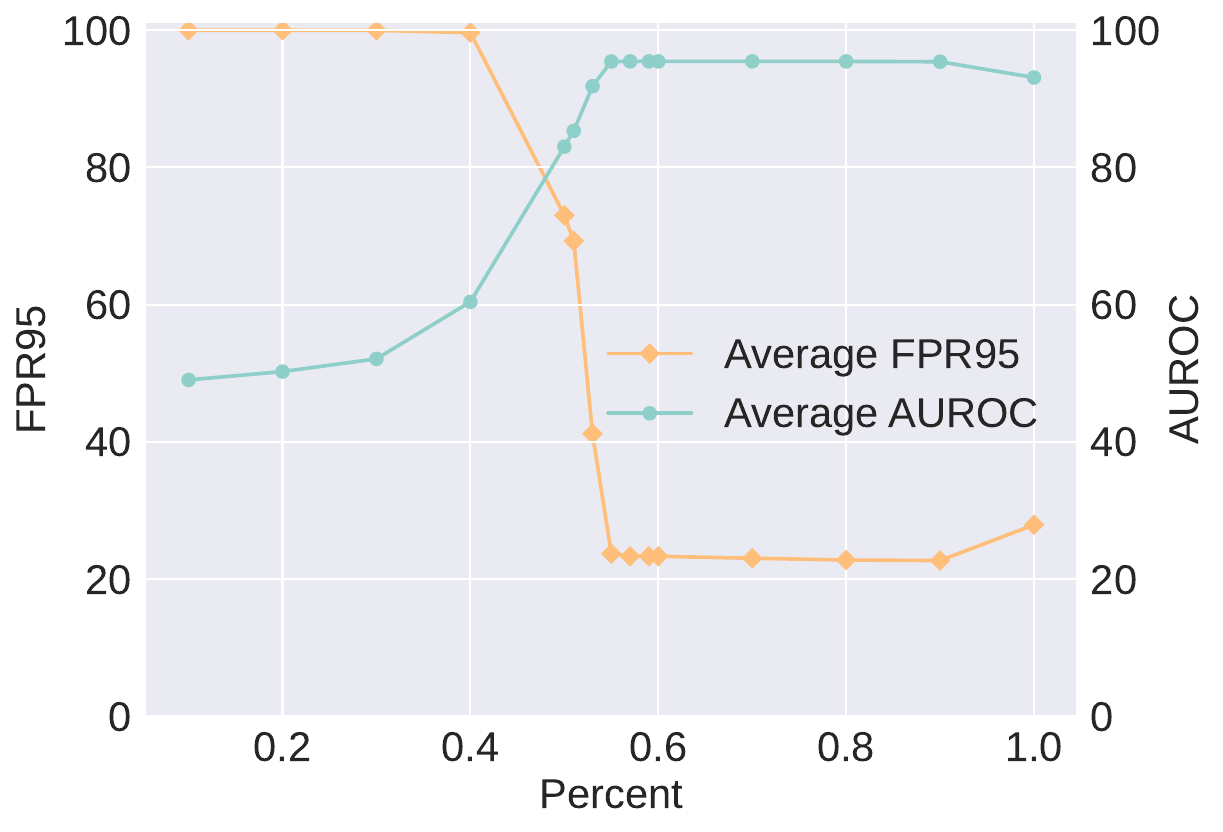}\label{2}}
  \vspace{0.3cm}
  \subfloat{\includegraphics[width=0.2\textwidth,height=0.15\textwidth]{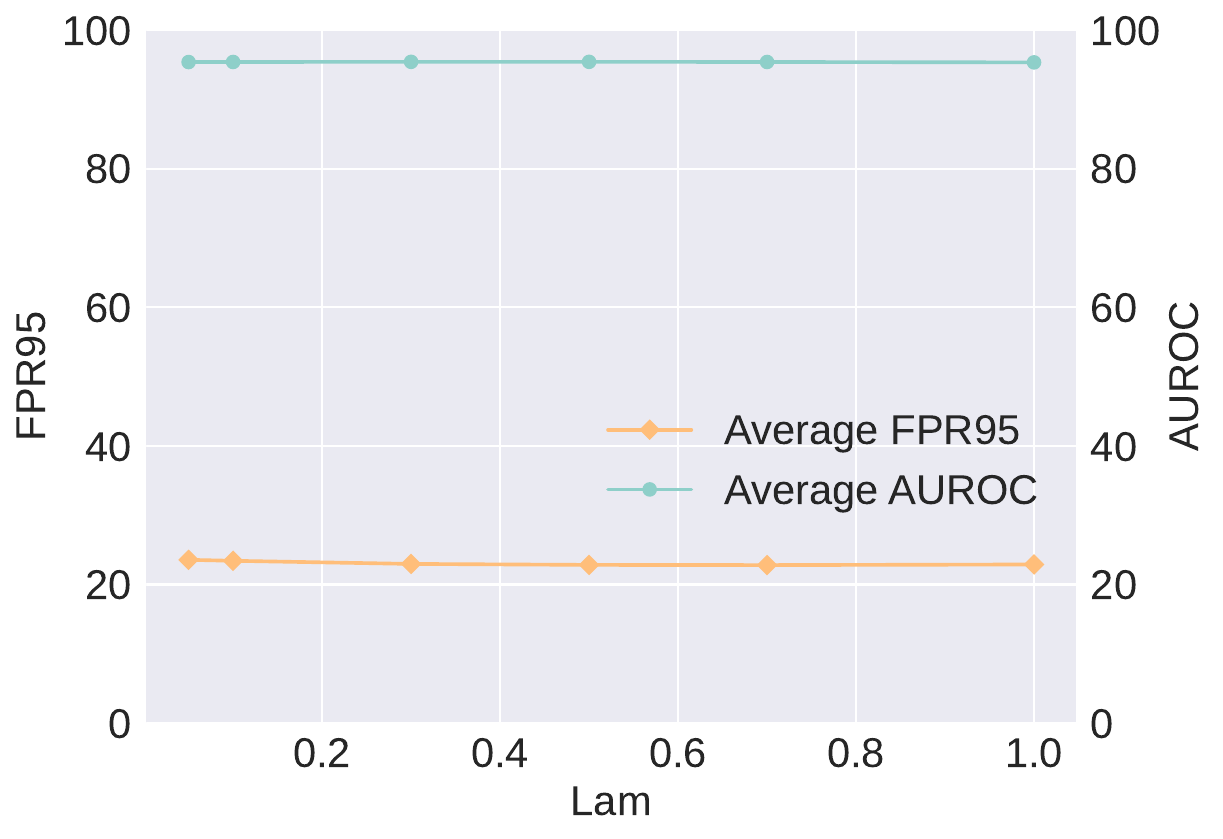}\label{3}}
  \hspace{0.5cm}
  \subfloat{\includegraphics[width=0.2\textwidth,height=0.15\textwidth]{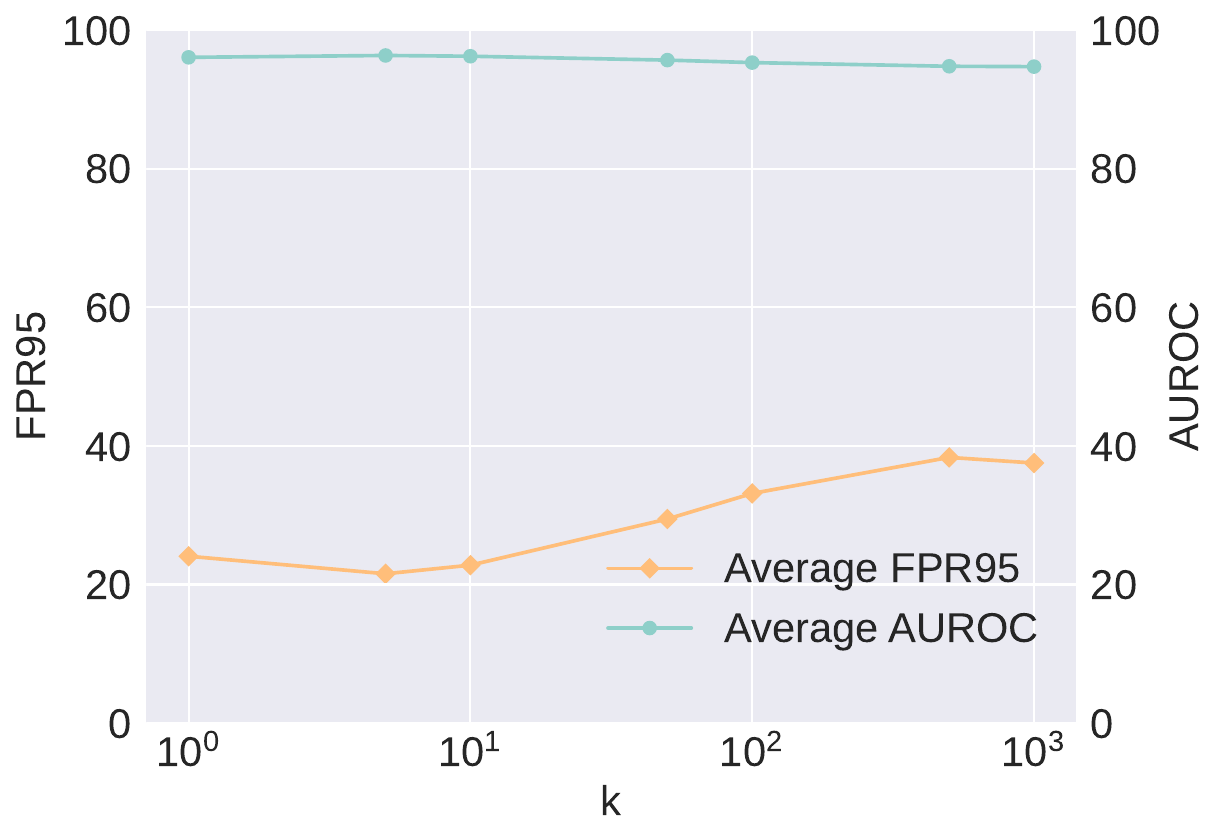}\label{4}}
  \caption{The influence of hyper-parameters in baseline methods on our detection performance. Top left: the temperature $T$ in Energy. Top right: the percent $p$ in ReAct. Bottom left: the tuning parameter $\lambda$ in BATS. Bottom right: the sort $k$ in KNN.}
  \label{effect_of_hyperparameter_in_other_methods}
  \vspace{-2.0ex}
\end{figure}

\textbf{Effect of Hyper-parameters in Baseline Methods.}
We analyze how the following hyper-parameters influence our detection performance on their corresponding methods:
\begin{itemize}
    \item The temperature $T$ in Energy \cite{energy}
    \item The percent $p$ in ReAct \cite{react}
    \item The tuning parameter $\lambda$ in BATS \cite{bats} 
    \item The sort $k$ in KNN \cite{knn}
\end{itemize}
Our experiments are conducted on ImageNet with a fixed reduced dimension of $K=1000$. The results are presented in Figure \ref{effect_of_hyperparameter_in_other_methods}. For the BATS and KNN methods, we observe that our performance remains stable with the change of hyper-parameters. In the case of the Energy method, the performance deteriorates with an increase in temperature, aligning with the observation in the original Energy \cite{energy} paper. Regarding the ReAct method, we observe a notable performance improvement when the percent changes from 0.5 to 0.6, which is expected since the gradients of ID data undergo a substantial impact when $p$ is small. These results provide valuable insights into the sensitivity and stability of our approach with respect to specific hyper-parameters in different baseline methods.

\vspace{-6mm}
\textcolor{black}{\section{Discussion}}
\textcolor{black}{
\subsection{Benefit of Dimensionality Reduction}
Beyond addressing the curse of dimensionality \cite{zimek2012survey}, in this part, we take a deeper look at the benefits of dimensionality reduction for OOD detection. Firstly, we evaluate the reconstruction error of gradients to analyze the influence of dimensionality reduction on ID and OOD data. The reconstruction error is defined as $\Vert G - GVV^T\Vert_2/\Vert G \Vert_2$, which reflects the ratio of the gradient component in the residual space to the whole gradient magnitude. Experiments are conducted on CIFAR10 benchmark with ResNet18 model, using SVHN \cite{svhn} as OOD dataset. The result is reported in Table \ref{1-3-2}, which reveals that gradients of ID data approximately cluster in the low-dimensional space while gradients of OOD data do not fall into it. This difference induces a larger distance between ID and OOD gradients in the low-dimensional space compared to the whole space, which is verified by our empirical results in Table \ref{1-3-1}. Therefore, the benefits of our method are not limited to utilizing complete gradient information, but also to widening the difference between ID and OOD gradients in the low-dimensional space.
}

\begin{table*}[htbp]\color{black}
  \centering
  \caption{Comparison of detection performance between using PCA and average gradient subspace on CIFAR10 benchmark with ResNet18. \label{2-2}}
  \begin{tabular*}{0.95\linewidth}{@{  }cccccccccccccc@{}}
    \toprule
    \multirow{2}{*}{Subspace} & \multirow{2}{*}{Dim} & \multicolumn{2}{c}{SVHN} & \multicolumn{2}{c}{Texture} & \multicolumn{2}{c}{LSUN} & \multicolumn{2}{c}{iSUN} & \multicolumn{2}{c}{Average}  \\
    & & FPR95$\downarrow$ & AUROC$\uparrow$ & FPR95$\downarrow$ & AUROC$\uparrow$ & FPR95$\downarrow$ & AUROC$\uparrow$ & FPR95$\downarrow$ & AUROC$\uparrow$ & FPR95$\downarrow$ & AUROC$\uparrow$\\
     \midrule  
     PCA & 200 & 12.72 & 97.51 & 32.20 & 94.62 & 13.36 & 97.41 & 27.91 & 95.79 & \textbf{21.55} & \textbf{96.33} \\
     PCA & 10 & 33.18 & 95.47 & 48.12 & 92.95 & 35.18 & 95.66 & 28.00 & 95.40 & 36.12 & 94.87 \\
     AG & 10 & 36.60 & 95.16 & 49.57 & 92.52 & 42.39 & 95.21 & 28.01 & 95.39 & 39.14 & 94.57 \\ 
  \bottomrule
\end{tabular*}
\end{table*}

\begin{table}[!t]\color{black}
    \centering
    \caption{Reconstruction error of gradients for Test (ID) and OOD data to the 200-dimensional principal subspace. The average reconstruction error of 1000 randomly sampled data is reported on each dataset. \label{1-3-2}}
    \begin{tabular}{c|c|c}
    \toprule
        Data & Test & OOD \\
        \midrule
        Reconstruction Error & 0.08 & 0.87 \\
    \bottomrule
    \end{tabular}
\end{table}

\begin{table}[!t]\color{black}
    \vspace{-2.0ex}
    \centering
    \caption{Nearest neighbor distance to training data gradients. The distance is calculated using the Euclidean distance between normalized gradients, as KNN method \cite{knn} does. The average nearest distance of 1000 randomly sampled data on Test and OOD datasets is respectively reported. A higher difference between Train-Test and Train-OOD indicates better detection performance.  
     \label{1-3-1}}
    \resizebox{\linewidth}{!}{
    \begin{tabular}{c|ccc|ccc}
    \toprule
        \multirow{2}{*}{Metric} & \multicolumn{3}{c|}{In whole space} & \multicolumn{3}{c}{In low-dimensional space} \\
         & Train-Test & Train-OOD & Diff & Train-Test & Train-OOD & Diff \\
        \midrule
        Distance & 1.32 & 1.87 & 0.55 & 0.49 & 1.84 & \textbf{1.35} \\
    \bottomrule
    \end{tabular}}
    \vspace{-3.0ex}
\end{table}

\vspace{-5mm}
\textcolor{black}{
\subsection{Comparison between PCA and Average Subspace}
In this part, we compare the detection performance of using PCA subspace and the alternative cheaper one (average gradient subspace, AG) on the CIFAR10 benchmark with ResNet18 model. The results are shown in Table \ref{2-2}, where we adopt the KNN \cite{knn} as our basic score function. From it, we can find that the 200-dimensional PCA subspace significantly outperforms the AG subspace. Meanwhile, the performance of the 10-dimensional PCA subspace slightly exceeds that of AG, consistent with our analysis in Sec \ref{Low-dimensional Gradient Extraction}. According to the result, using PCA subspace for dimensionality reduction is a better choice. But in practice, we can flexibly choose the subspace considering the trade-off between detection performance and computational overhead, since our analysis in Sec \ref{Time Overhead} reveals that computing the average gradient only requires little time overhead.
}

\textcolor{black}{\subsection{Impact of Outlier Exposure}}
\noindent
\textcolor{black}{
Our previous experiments have verified that in vanilla training, gradient provides a more significant difference between ID and OOD data compared to output/feature information. To study the universal applicability of our approach, we explore the influence of outlier exposure (OE, \cite{oe}), a model fine-tuning method using auxiliary OOD data, on the gradients. We follow the common training setting in OE \cite{oe} to fine-tune a vanilla ResNet18 model with CIFAR10 \cite{cifar10} as ID dataset and Tiny Images \cite{torralba200880} as auxiliary OOD Dataset. Specific settings are in Appendix C. Based on the OE-trained model, we evaluate the detection performance of common post-hoc scores including MSP \cite{msp}, Energy \cite{energy}, ReAct \cite{react}, BATS \cite{bats} and KNN \cite{knn} on the forward feature, and our detection performance based on the low-dimensional gradient with MSP score. The first four scores \cite{msp,energy,react,bats} are calculated based on model outputs, and the KNN score \cite{knn} is computed using the k-th nearest distance defined on feature space. Experiment results are reported in Table \ref{1-1}, from which we can see that OE makes the difference between ID and OOD more significant for both forward and backward information. Notice here we directly used the existing OE method that is designed for output-based scores, which hence explicitly enlarges the discrepancy of output space. In this case, our low-dimensional gradient achieves comparable performance, fully demonstrating the versatility of our approach. We believe that there is potential to further improve our detection performance by designing a suitable training loss using auxiliary OOD data to explicitly enlarge the discrepancy between ID and OOD gradients.
}
\begin{table}[htbp]\color{black}
  \centering
  \caption{Results based on an OE-trained model on CIFAR10 with ResNet18. The best result is in bold. \label{1-1}}
  \resizebox{\linewidth}{!}{
  \begin{tabular}{cccccccc}
    \toprule
    Methods & MSP & Energy & ReAct & BATS & KNN & Ours \\
    \midrule
    FPR95$\downarrow$ & 2.46 & \textbf{2.35} & 2.35 & 4.06 & 8.60 & 2.36 \\
    AUROC$\uparrow$ & \textbf{98.57} & 98.55 & 98.55 & 98.40 & 94.73 & 97.37 \\
  \bottomrule
\end{tabular}}
\vspace{-3.0ex}
\end{table}

\textcolor{black}{\subsection{Time Overhead}\label{Time Overhead}}
\textcolor{black}{
In this part, we report the specific time cost of our approach. All the experiments are conducted on four NVIDIA GeForce RTX 2080Ti GPUs. 
Generally, our method consists of two stages: offline subspace extraction and online score calculation. The offline stage can be pre-computed and saved using the in-distribution (ID) training dataset, meaning it only needs to be performed once.
The results presented in Table~\ref{1-4-1} and Table~\ref{1-4-2} exhibit the computation time of each stage. We can observe that the PCA-based subspace extraction takes a longer time compared to the average gradient approach~(Table~\ref{1-4-1}). However, as to the inference cost, the computation time of our method is almost equal to the time required for a single backward propagation as shown in Table~\ref{1-4-2}, which is considered acceptable for practical applications.
}
\begin{table}[htbp]\color{black}
    \centering
    \caption{Computation time of the subspace extraction stage. AVG represents the subspace spanned by average gradients and PCA denotes the subspace spanned by principal components.}
    \label{1-4-1}
    \resizebox{\linewidth}{!}{
    \begin{tabular}{cccccc}
    \toprule
        Dataset & Architecture & Param & Subspace Type & Dim & Time(h) \\
        \midrule
        \multirow{2}{*}{CIFAR10} & \multirow{2}{*}{ResNet18} & \multirow{2}{*}{11.17M} & AVG & 10 & 0.006 \\ 
         & & & PCA & 200 & 14.04 \\
         \midrule
         ImageNet & ResNet50 & 25.56M & AVG & 1000 & 0.75 \\
    \bottomrule
    \end{tabular}}
\end{table}

\begin{table}[htbp]\color{black}
    \centering
    \caption{Computation time of the detecting stage, which is consisted of calculating backward gradients, projecting gradients into the low-dimensional subspace and predicting outputs by the linear network in Eq (\ref{linear network}). We report the detection time of one image.}
    \label{1-4-2}
    \resizebox{\linewidth}{!}{
    \begin{tabular}{ccccccc}
    \toprule
        \multirow{2}{*}{Architecture} & \multirow{2}{*}{Param} & \multirow{2}{*}{Dim} & \multicolumn{3}{c}{Time(ms)} & \multirow{2}{*}{Overall(ms)} \\
         & & & Backward & Projection & Prediction & \\
         \midrule
        \multirow{2}{*}{ResNet18} & \multirow{2}{*}{11.17M} & 10 & $686.01$ & $0.23$ & $0.63$ & $686.87$\\
         & & 200 & $686.01$ & $0.26$ & $0.64$ & $686.91$\\
         \midrule
         \multirow{1}{*}{ResNet50} & \multirow{1}{*}{25.56M} & 1000 & 747.22 & 0.32 & 0.70 & 748.24 \\
    \bottomrule
    \end{tabular}} 
\end{table}

\section{Limitation}\label{limitation}
\textcolor{black}{Our method involves certain computational overhead for both the offline subspace extraction and online score calculation stages.} In practice, it is necessary to consider the available computing resources and optimize the implementation for efficient execution of these processes.
\textcolor{black}{
One possible way to accelerate our subspace extraction is by random sampling \cite{tsuyuzaki2020benchmarking,abraham2014fast}, \textit{i.e.}, employing a subset of the training data to calculate the principal components.}
\textcolor{black}{
Moreover, while our results are promising, there are some cases where feature-based methods may yield better performance compared to gradient-based ones. As the primary focus of this paper is to investigate the feasibility of utilizing gradient embedding, we leave further improvements and comparative studies for future research.
}

\section{Conclusion}
This paper presents a pioneering study exploring the utilization of complete parameter gradient information for OOD detection. To solve the problem arising from the high dimensionality of gradients, we propose to conduct dimension reduction on gradients using our designated subspace, which comprises the principal components of gradients. With the low-dimensional representations of gradients, we subsequently explore their integration with various detection algorithms. Our extensive experiments demonstrate that our low-dimensional gradients can notably improve performance across a wide range of detection tasks. We hope our work inspires further research on leveraging parameter gradients for OOD detection.

\textcolor{black}{
\section*{Acknowledgment}
The authors would like to thank the anonymous reviewers for their insightful comments and valuable suggestions.}\\


\bibliographystyle{IEEEtran}
\bibliography{reference.bib}

\newpage
\appendix

\section{Efficient PCA for DNNs}
Let us denote the gradient over the training dataset as $G\in \mathcal{R}^{n\times |\theta|}$, where $n$ denotes the sample number and $|\theta|$ denotes the parameter number. To extract the low-dimensional subspace where the gradient principal components reside, we need to compute the top-K eigenvector of the gradient covariance $C=G^TG$. The main difficulties lie in two aspects: 1. The gradient matrix $G$ is too large to calculate and save since both $n$ and $|\theta|$ are enormous for modern DNNs. 2. Even if $G$ and $C$ are obtained, the eigen decomposition of $C$ cannot be directly calculated because of its high dimensionality.

To solve the above problems, we employ the power iteration method \cite{bathe1971solution,golub2000eigenvalue} to efficiently calculate the top-$K$ eigenvectors of $C$, while ensuring computational feasibility. Noticing that the covariance matrix $C$ is a dot product of $G^T$ and $G$, each iteration step of the power method can be decomposed into two steps, as shown in Algorithm \ref{Power Iteration Method}. And for each decomposed step, there exist efficient implementations in Pytorch \cite{pytorch} called Jacobian Vector Product (JVP) and Vector Jacobian Product (VJP), respectively, calculated at the same order of computational costs of one vanilla backward-pass and forward-pass of DNNs. Specific algorithms are shown in Algorithm \ref{1} and Algorithm \ref{2}, where we integrate dataloaders into the algorithm. With this practical subspace extraction algorithm, we can obtain a low-dimensional subspace denoted by $v\in\mathcal{R}^{|\theta|\times K}$ that encompasses the principal components of gradients.
\begin{algorithm}[ht]
\caption{Power Iteration Method}\label{Power Iteration Method}
\begin{algorithmic}[1]
\REQUIRE covariance matrix $C=G^TG \in R^{|\theta| \times |\theta|}$, iteration step $T$, dimension number $K$
\ENSURE top-$K$ eigenvectors of $C$
\STATE $ \textbf{Draw random vector } v\in R^{|\theta|\times K}  $
\STATE $ \textbf{Schmidt orthogonalization of } v $
\FOR{$i=1$ to $T$}
\STATE $ v \gets Gv$ \COMMENT{$v=Gv \in R^{n\times K}$}
\STATE $ v \gets G^Tv$ \COMMENT{$v=G^Tv\in R^{|\theta|\times K}$}
\STATE $\textbf{Schmidt orthogonalization of } v $
\ENDFOR
\STATE \textbf{return}  $v$
\end{algorithmic}
\end{algorithm}

\begin{algorithm}[ht]
\caption{Efficient Calculation of $Gv$}\label{1}
\begin{algorithmic}[1]
\REQUIRE vector $v\in R^{|\theta|\times K}$, model $f_\theta$, training dataloader $D$, mean matrix $\mathcal{M}$, variance matrix $\mathcal{I}$
\ENSURE $Gv$
\STATE $v \gets (diag(\mathcal{I})^{-\frac{1}{2}}) v$
\STATE $v_{out} \gets [\ ]$
\FOR{$x$ in $D$}
    \STATE $ a \gets$ \texttt{functorch.jvp}$(f_\theta, x, v) - \mathcal{M}v$\\ 
    \COMMENT{$a\in R^{b\times K}, b:$ batch size}
    \STATE $ v_{out} \gets [a, v_{out}]$
\ENDFOR
\STATE \textbf{return}  $v_{out}$
\end{algorithmic}
\end{algorithm}

\begin{algorithm}[ht]
\caption{Efficient Calculation of $G^Tv$}\label{2}
\begin{algorithmic}[1]
\REQUIRE vector $v\in R^{n\times K}$, model $f_\theta$, training dataloader $D$, mean matrix $\mathcal{M}$, variance matrix $\mathcal{I}$
\ENSURE $G^Tv$
\STATE $v_{out} \gets 0,\ j \gets 0$
\FOR{$x$ in $D$}
    \STATE $ v_t \gets v[j:j+b]$ \COMMENT{$v_t\in R^{b\times K}, b:$ batch size}
    \STATE $ a \gets$ \texttt{functorch.vjp}$(f_\theta, x, v_t)$
    \STATE $ v_{out} \gets v_{out}+a $ 
    \STATE $ j \gets j+b$
\ENDFOR
\STATE $ s \gets \{\sum_i v[i,j]\}$ \COMMENT{$s\in R^{K}$}
\STATE $ \text{Expand the dimension of } s \text{ to } R^{|\theta|\times K}$
\STATE $ v_{out} \gets (diag(\mathcal{I})^{-\frac{1}{2}})(v_{out} - \mathcal{M}s)$
\STATE \textbf{return}  $v_{out}$
\end{algorithmic}
\end{algorithm}

\begin{figure}[htbp]
  \centering
  \subfloat[]{\includegraphics[width=0.20\textwidth]{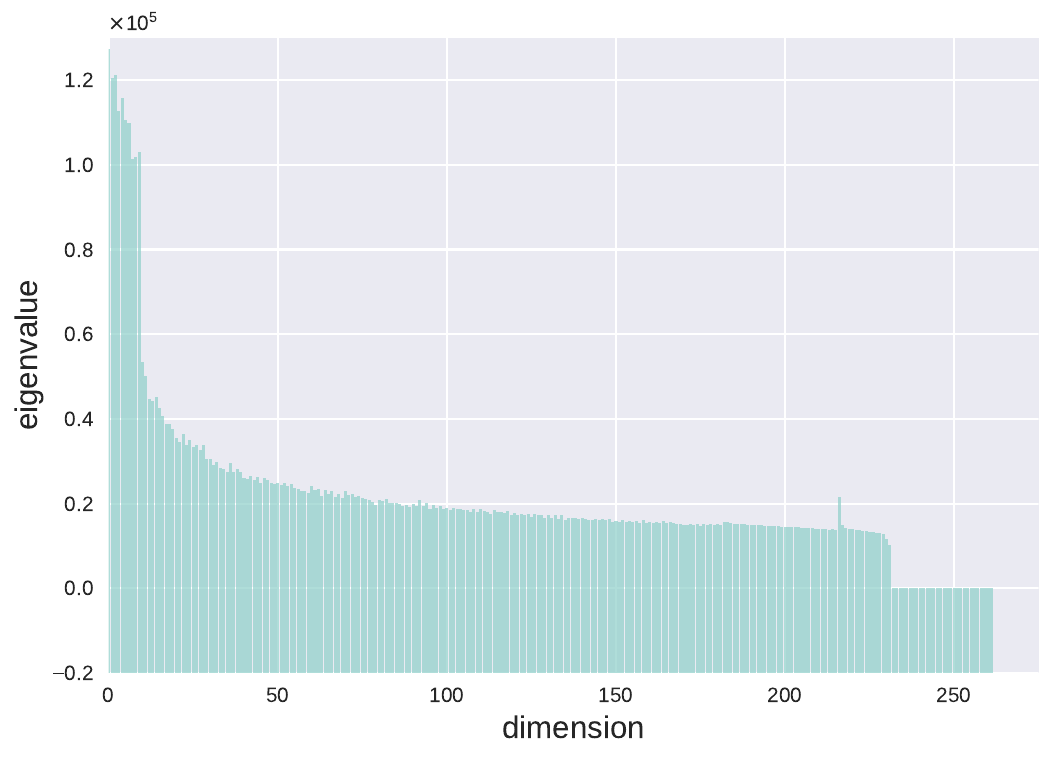}\label{his}}
  \hspace{0.5cm}
  \subfloat[]{\includegraphics[width=0.20\textwidth]{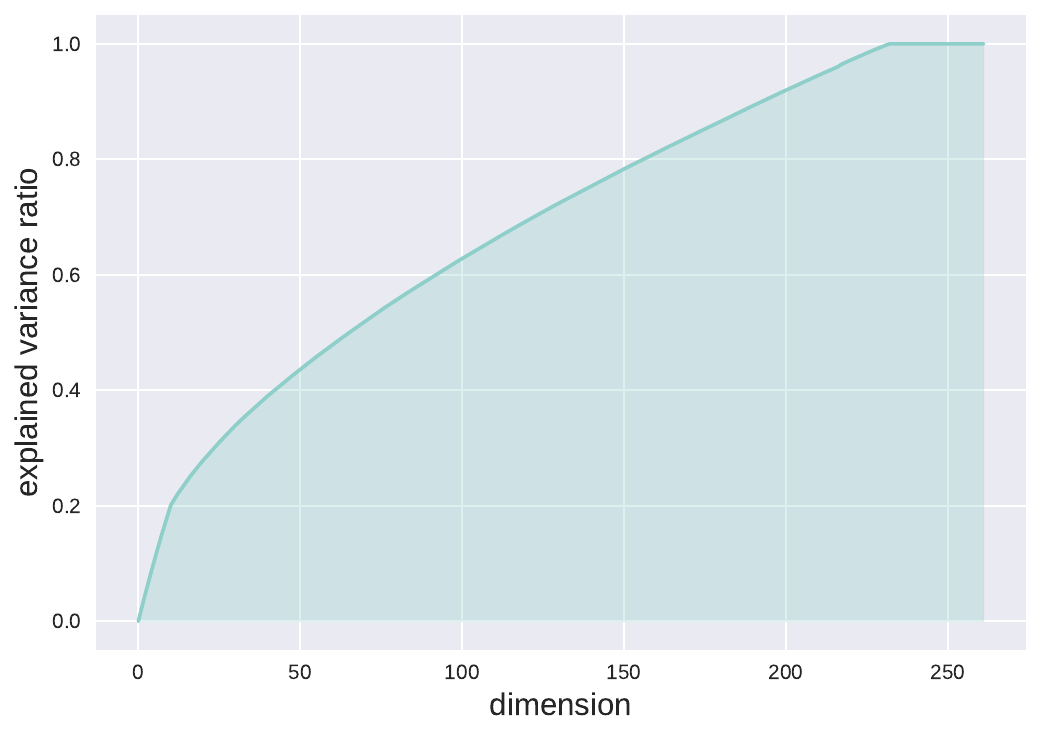}\label{ratio}}
  \caption{(a): The magnitude of eigenvalues of the gradient covariance matrix; (b): The explained variance ratio as a function of the number of components. The results are derived from a well-trained ResNet18 model on CIFAR10.}
  \label{fig:side-by-side}
\end{figure}

Figure \ref{his} shows the magnitude of the eigenvalues of the gradient covariance matrix calculated by our algorithm. It can be observed that the eigenvalues tend to be concentrated within the initial 200 dimensions, with subsequent eigenvalues abruptly diminishing to zero. Additionally, we present Figure \ref{ratio}, which illustrates the explained variance ratio as a function of the number of components. The results reveal that the top-200 principal components account for approximately 90$\%$ of the total variance. The above findings provide empirical validation for the low-dimensional nature of gradients.

\section{Gradient Directional Similarity}
We examine the cosine similarity between the parameter gradient of a sample and its corresponding class-mean gradient on both the CIFAR10 and ImageNet datasets. In a high-dimensional space, the cosine similarity between two random vectors tends to approach zero, as verified in \cite{zimek2012survey}. However, the average cosine similarity values presented in Table \ref{cosine similarity} are significantly higher than zero, which implies that gradients of samples belonging to the same class tend to align in similar directions.

\begin{table}[ht]
    \centering
    \caption{The cosine similarity between the parameter gradient of a sample and its corresponding class-mean gradient. We report the average value, respectively, on the training and test datasets. \label{cosine similarity}}
    \begin{tabular}{ccc}
    \toprule
        Dataset & Training Set & Test Set \\
         \midrule
        CIFAR10 & 0.59 & 0.47 \\
        ImageNet & 0.51 & 0.45\\
    \bottomrule
    \end{tabular}
\end{table}

\section{Training Setting for Outlier Exposure}
The outlier exposure method \cite{oe}, which utilizes auxiliary OOD data to fine-tune the model, has shown excellent performance in OOD detection. It can be formalize as the following optimization problem:
\begin{equation}\label{OE problem}
   \min_f \quad \mathbb{E}_{(x,y)\sim D_{in}} L_{CE}(x,y)+\lambda \mathbb{E}_{x\sim D_{out}^{aux}} L_{OE}(x)
\end{equation}
where $f$ is the model hypothesis, $L_{CE}$ is the cross-entropy loss on ID data, and $L_{OE}$ is the outlier exposure loss on auxiliary OOD data defined as the cross-entropy between OOD outputs and uniformly distributed labels for classification problem. The $\lambda$ is a balance hyper-parameter, usually set to $0.5$ in experiments. In the following, we present the detailed experimental settings.\\
\textbf{OOD Datasets.} We randomly choose 300K samples from the 80 Million Tiny Images \cite{torralba200880} as our auxiliary OOD dataset. \\
\textbf{Pre-training Setups.} We employ ResNet18 \cite{he2016deep} trained for 200 epochs, with batch size 128, init learning rate 0.1, momentum 0.9, weight decay 0.0005, and cosine schedule. \\
\textbf{Fine-tuning Setups.} We adopt the model parameter of the 99th epoch in the pre-training process as our initial network parameters, and then add auxiliary OOD data to train the model for 50 epochs with ID batch size 128, OOD batch size 256, initial learning rate 0.07, momentum 0.9, weight decay 0.0005 and cosine schedule.



\end{document}